\documentclass{article}

\usepackage[utf8]{inputenc}
\usepackage[T1]{fontenc}

\usepackage{color,xcolor}
\usepackage{epsfig}
\usepackage{graphicx}
\usepackage{duckuments}

\usepackage{adjustbox}
\usepackage{array}
\usepackage{booktabs}
\usepackage{colortbl}
\usepackage{float,wrapfig}
\usepackage{hhline}
\usepackage{multirow}
\usepackage{subcaption}
\usepackage[font=small]{caption}
\usepackage{makecell}
\usepackage{listings}

\usepackage{amsmath,amsfonts,amsthm,amssymb}
\usepackage{bm}
\usepackage{nicefrac}
\usepackage{microtype}

\usepackage{changepage}
\usepackage{extramarks}
\usepackage{fancyhdr}
\usepackage{lastpage}
\usepackage{setspace}
\usepackage{soul}
\usepackage{xspace}
\usepackage{indentfirst}
\usepackage{pifont}
\usepackage{cuted}
\usepackage{wrapfig}
\definecolor{cvprblue}{rgb}{0.21,0.49,0.74}

\usepackage[pagebackref,breaklinks,colorlinks,citecolor=cvprblue]{hyperref}
\usepackage{url}

\usepackage{algorithm, algorithmic}
\usepackage{enumitem}

\usepackage{wasysym}
\usepackage{todonotes}
\usepackage{pifont}
\usepackage{fancyvrb}
\usepackage{fvextra}
\newcolumntype{L}[1]{>{\raggedright\let\newline\\\arraybackslash\hspace{0pt}}m{#1}}
\newcolumntype{C}[1]{>{\centering\let\newline\\\arraybackslash\hspace{0pt}}m{#1}}
\newcolumntype{R}[1]{>{\raggedleft\let\newline\\\arraybackslash\hspace{0pt}}m{#1}}

\newcommand{\sect}[1]{\S~\ref{sect:#1}}

\newcommand{\eqn}[1]{Equation~\ref{eq:#1}}

\newcommand{\fig}[1]{Figure~\ref{fig:#1}}

\newcommand{\tbl}[1]{Table~\ref{tab:#1}}

\newcommand{\lblsect}[1]{\label{sect:#1}}

\newcommand{\ignorethis}[1]{}

\makeatletter
\DeclareRobustCommand\onedot{\futurelet\@let@token\@onedot}
\def\@onedot{\ifx\@let@token.\else.\null\fi\xspace}

\makeatother

\definecolor{citecolor}{rgb}{34,139,34}
\definecolor{mydarkblue}{rgb}{0,0.08,1}
\definecolor{mydarkgreen}{rgb}{0.12,0.7,0.12}
\definecolor{mydarkred}{rgb}{0.8,0.02,0.02}
\definecolor{mydarkorange}{rgb}{0.40,0.2,0.02}
\definecolor{mypurple}{RGB}{111,0,255}
\definecolor{myred}{rgb}{1.0,0.0,0.0}
\definecolor{mygold}{rgb}{0.75,0.6,0.12}
\definecolor{mydarkgray}{rgb}{0.66,0.66,0.66}

\definecolor{darkgreen}{rgb}{0.15, 0.75, 0.15}
\definecolor{mitblue}{rgb}{0.88,0.95,0.96}
\definecolor{lightblue}{rgb}{0.90, 0.95, 0.99}

\newcommand{\QUEST}{QUEST\xspace}

\newcommand{\ours}{\textsc{LoSA}\xspace}
\newcommand{\kvinflation}{KV Inflation\xspace}
\newcommand{\localityrepchanges}{locality of representation changes\xspace}
\newcommand{\LocalityRepChanges}{Locality of Representation Changes\xspace}
\newcommand{\activetokens}{active tokens\xspace}
\newcommand{\Activetokens}{Active tokens\xspace}
\newcommand{\stabletokens}{stable tokens\xspace}
\newcommand{\Stabletokens}{Stable tokens\xspace}

\usepackage{amsmath,amsfonts,bm}

\def\eqref#1{equation~\ref{#1}}

\def\1{\bm{1}}

\def\vo{{\bm{o}}}

\def\vq{{\bm{q}}}

\def\vs{{\bm{s}}}

\def\vx{{\bm{x}}}

\def\mK{{\bm{K}}}

\def\mO{{\bm{O}}}

\def\mQ{{\bm{Q}}}

\def\mV{{\bm{V}}}

\def\mX{{\bm{X}}}

\DeclareMathAlphabet{\mathsfit}{\encodingdefault}{\sfdefault}{m}{sl}
\SetMathAlphabet{\mathsfit}{bold}{\encodingdefault}{\sfdefault}{bx}{n}

\def\gS{{\mathcal{S}}}

\def\sR{{\mathbb{R}}}

\newcommand{\R}{\mathbb{R}}

\newcommand{\softmax}{\mathrm{softmax}}

\ifx\theorem\undefined
\newtheorem{theorem}{Theorem}[section]
\fi

\ifx\example\undefined
\newtheorem{example}{Example}[section]
\fi

\ifx\property\undefined
\newtheorem{property}{Property}
\fi

\ifx\lemma\undefined
\newtheorem{lemma}{Lemma}[section]
\fi

\ifx\proposition\undefined
\newtheorem{proposition}{Proposition}[section]
\fi

\ifx\remark\undefined
\newtheorem{remark}{Remark}
\fi

\ifx\corollary\undefined
\newtheorem{corollary}{Corollary}[section]
\fi

\ifx\definition\undefined
\newtheorem{definition}{Definition}[section]
\fi

\ifx\conjecture\undefined
\newtheorem{conjecture}{Conjecture}[section]
\fi

\ifx\axiom\undefined
\newtheorem{axiom}[theorem]{Axiom}
\fi

\ifx\claim\undefined
\newtheorem{claim}[theorem]{Claim}
\fi

\ifx\assumption\undefined
\newtheorem{assumption}{Assumption}
\fi

\ifx\problem\undefined
\newtheorem{problem}{Problem}[section]
\fi

\ifx\fact\undefined
\newtheorem{fact}{Fact}
\fi

\usepackage{xcolor}

\usepackage[preprint]{icml2026}

\icmltitlerunning{LoSA: Locality Aware Sparse Attention for Block-Wise Diffusion Language Models}

\begin{document}

\twocolumn[
  \icmltitle{LoSA: Locality Aware Sparse Attention for Block-Wise Diffusion Language Models}

  \icmlsetsymbol{equal}{*}

  \begin{icmlauthorlist}
    \icmlauthor{Haocheng Xi}{ucb}
    \icmlauthor{Harman Singh}{ucb}
    \icmlauthor{Yuezhou Hu}{ucb}
    \icmlauthor{Coleman Hooper}{ucb}
    \icmlauthor{Rishabh Tiwari}{ucb}
    \icmlauthor{Aditya Tomar}{ucb}
    \icmlauthor{Minjae Lee}{furiosa}
    \icmlauthor{Wonjun Kang}{furiosa}
    \icmlauthor{Michael Mahoney}{ucb,icsi,lbnl}
    \icmlauthor{Chenfeng Xu}{ut}
    \icmlauthor{Kurt Keutzer}{ucb}
    \icmlauthor{Amir Gholami}{ucb}
  \end{icmlauthorlist}

  \icmlaffiliation{ucb}{University of California, Berkeley}
  \icmlaffiliation{furiosa}{FuriosaAI}
  \icmlaffiliation{icsi}{ICSI}
  \icmlaffiliation{lbnl}{LBNL}
  \icmlaffiliation{ut}{UT Austin}

  \icmlcorrespondingauthor{Amir Gholami}{amirgh@berkeley.edu}

  \icmlkeywords{Machine Learning, ICML}

  \vskip 0.3in
]

\printAffiliationsAndNotice{}

\begin{abstract}
Block-wise diffusion language models (DLMs) generate multiple tokens in any order, offering a promising alternative to the autoregressive decoding pipeline. However, they still remain bottlenecked by memory-bound attention in long-context scenarios. Na\"{i}ve sparse attention fails on DLMs due to a \kvinflation{} problem, where different queries select different prefix positions, making the union of accessed KV pages large.
To address this, we observe that between consecutive denoising steps, only a small fraction of \activetokens{} exhibit significant hidden-state changes, while the majority of \stabletokens{} remain nearly constant.
Based on this insight, we propose \ours{} (\textbf{Lo}cality-aware \textbf{S}parse \textbf{A}ttention), which reuses cached prefix-attention results for stable tokens and applies sparse attention only to active tokens. This substantially shrinks the number of KV indices that must be loaded, yielding both higher speedup and higher accuracy.
Across multiple block-wise DLMs and benchmarks, \ours{} preserves near-dense accuracy while significantly improving efficiency, achieving up to $+9$ points in average accuracy at aggressive sparsity levels while maintaining $1.54\times$ lower attention density. It also achieves up to $4.14\times$ attention speedup on RTX A6000 GPUs, demonstrating the effectiveness of the proposed method.
\end{abstract}

\begin{figure}[t!]
\centering
\includegraphics[width=0.8\linewidth]{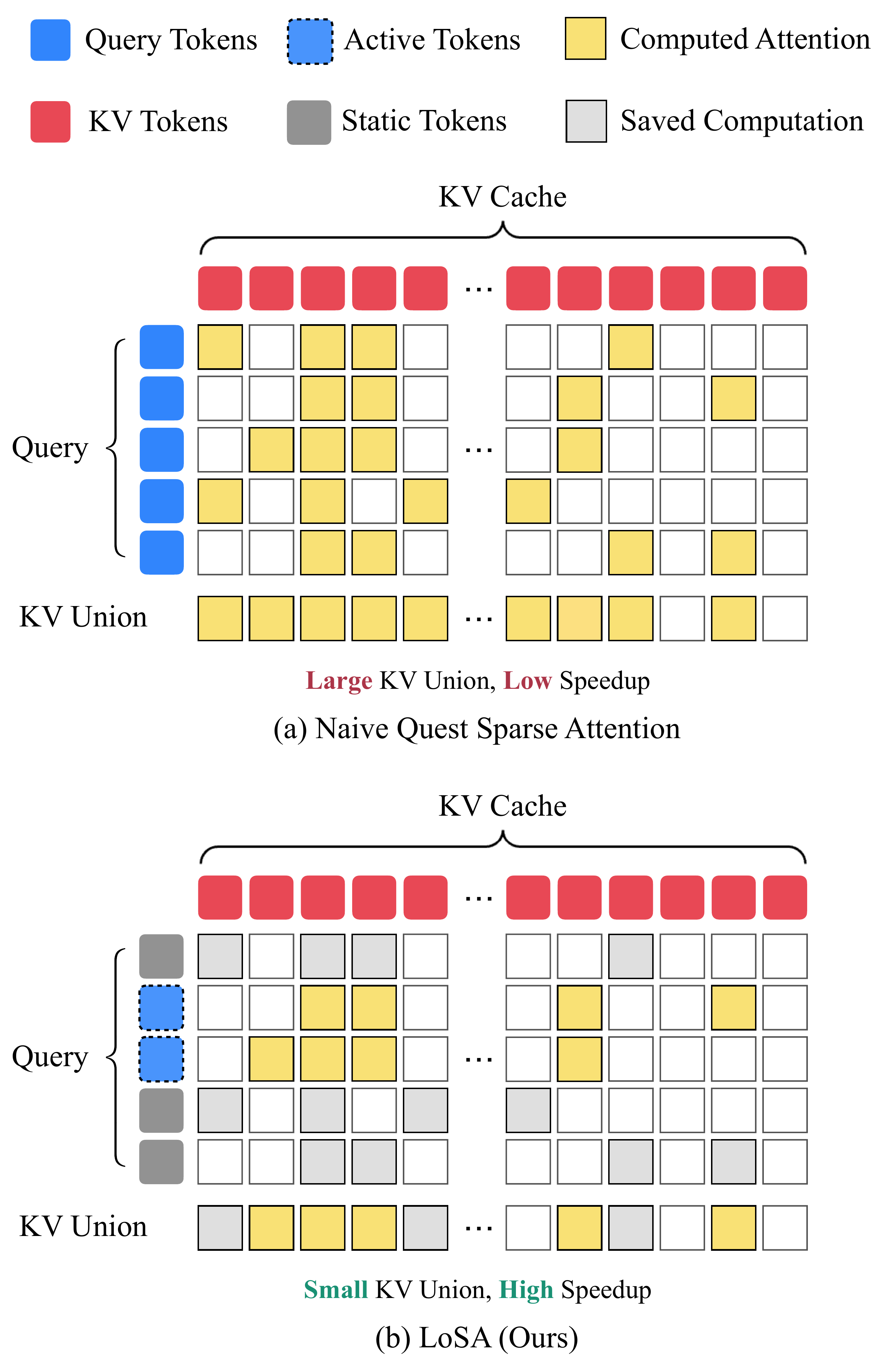}
\caption{
Illustration of the KV-union effect: in block diffusion, each query selects a small set of prefix KV positions, but the effective cost is determined by the union across the block, inflating KV access.
Our method computes sparse attention only for selected (active) tokens during denoising and reuses cached results for the others (static tokens), substantially reducing the size of the KV union and the latency.
}
\label{fig:KVInflationTeaser}
\end{figure}

\section{Introduction}

\begin{figure*}[h]
\centering
\begin{minipage}[t]{0.32\textwidth}
  \hspace{0.5cm}{Timestep $t-1$: $\|q_{t-1}\|_2$}\\
  \includegraphics[
    width=\linewidth,
    trim = 0mm 0mm 0mm 0mm,
    clip
  ]{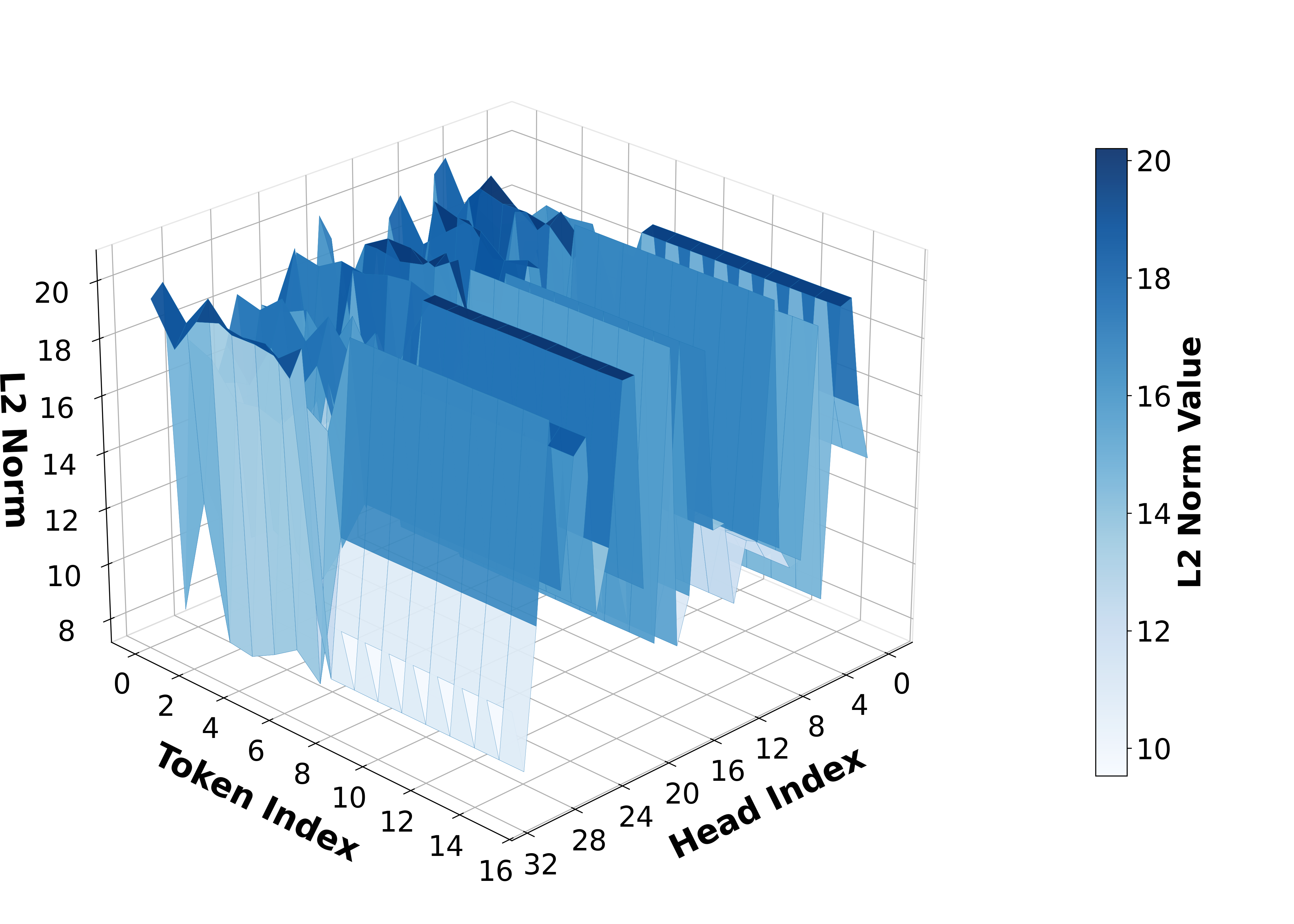}
\end{minipage}\hfill
\begin{minipage}[t]{0.32\textwidth}
  \hspace{0.8cm}{Timestep $t$: $\|q_{t}\|_2$}\\
  \includegraphics[
    width=\linewidth,
    trim = 0mm 0mm 0mm 0mm,
    clip
  ]{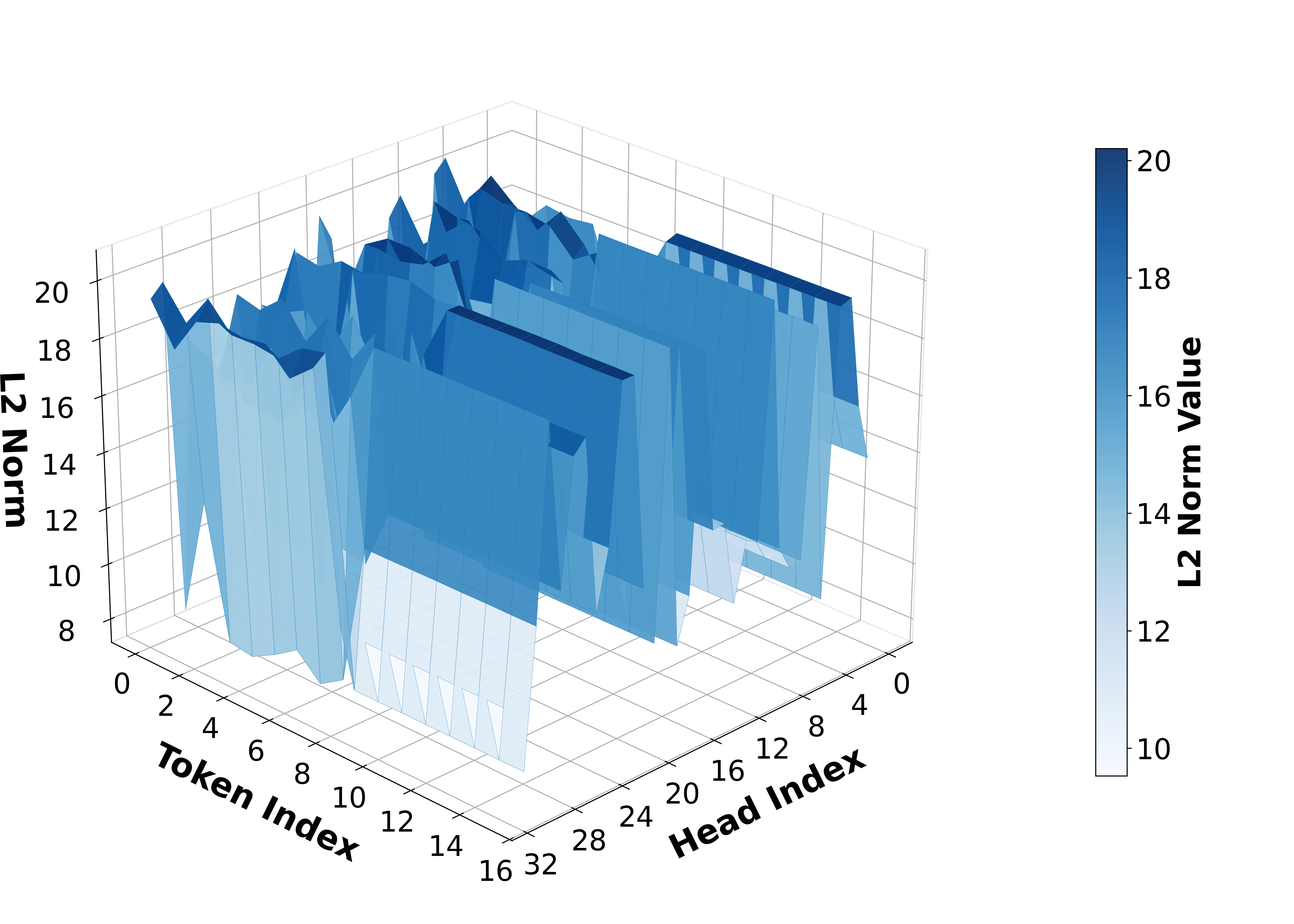}
\end{minipage}\hfill
\begin{minipage}[t]{0.32\textwidth}
  \hspace{0.10cm}{Timestep $t$: MSE($q_{t-1}, q_t$)}\\
  \includegraphics[
    width=\linewidth,
    trim = 0mm 0mm 0mm 0mm,
    clip
  ]{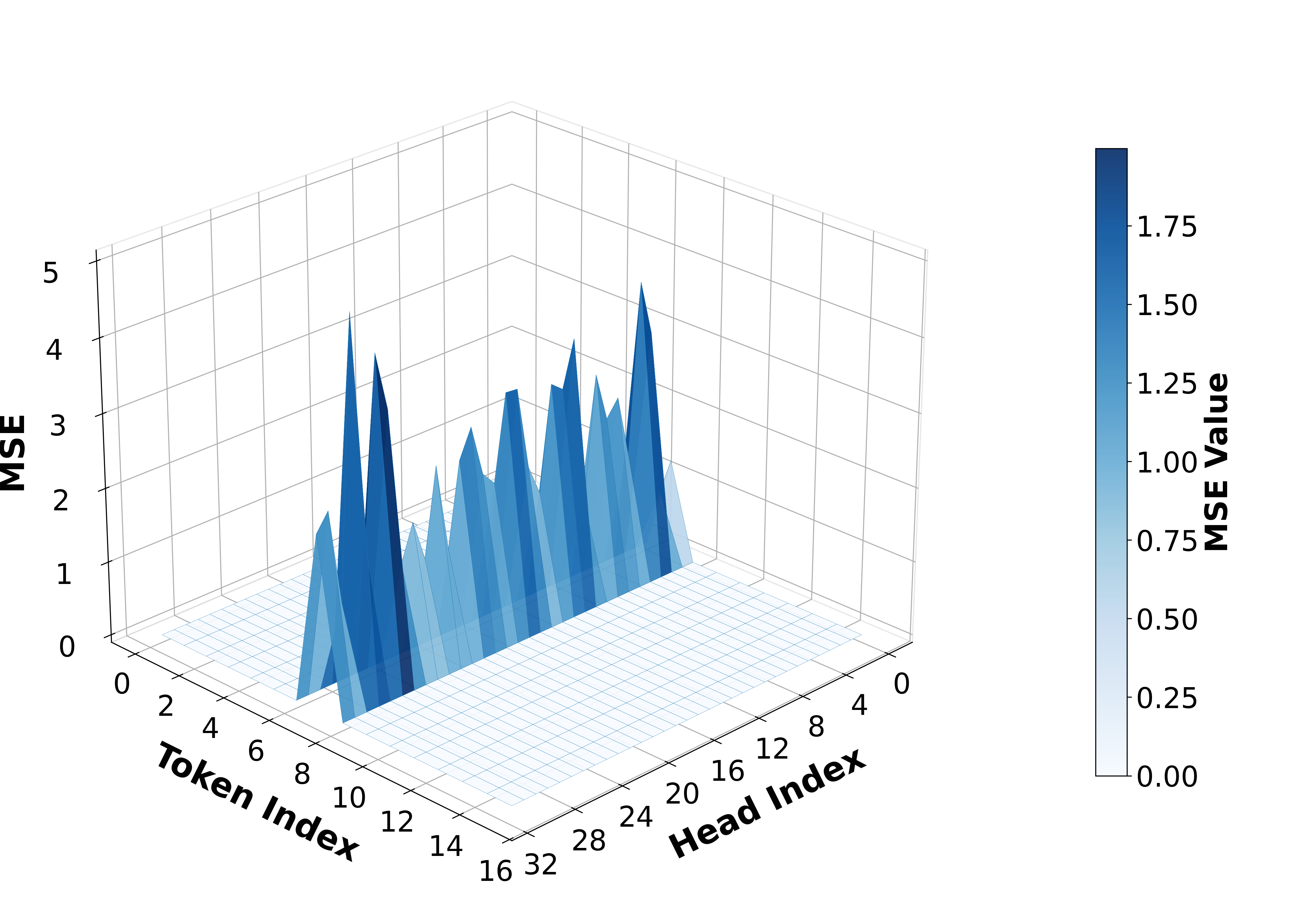}
\end{minipage}
\caption{Visualizing representation change locality across denoising steps. Left and middle plots show the L2 norm of query vectors at timesteps $t-1$ and $t$ respectively. The right plot shows the per-token MSE between query vectors across the two timesteps. Only a small fraction of tokens exhibit large changes, motivating the reuse of cached-prefix attention for stable tokens.}
\label{fig:locality_intro}
\end{figure*}

Diffusion language models (DLMs) have emerged as a compelling alternative to autoregressive transformers for text generation~\citep{austin2021structured, sahoo2024simple, lou2024discrete, nie2025large}. Unlike autoregressive models that generate tokens sequentially, DLMs generate tokens via iterative denoising and can update a block of tokens in any order, offering potential benefits for complex reasoning tasks. Recent block-wise DLMs~\citep{arriola2025block, cheng2025sdar, ye2025dream} produce tokens in diffusion-based blocks, enabling any-order, non-autoregressive generation within a block while preserving autoregressive dependencies across blocks. These models rely on a KV cache to reuse prefix representations.
Despite this flexibility, they remain bottlenecked by prefix attention: at each denoising step, all tokens in the block must attend to the entire cached prefix. For long contexts, the cost of repeatedly loading the KV cache dominates inference latency~\citep{dao2022flashattention}.

A natural approach to accelerate block-wise DLMs is to apply sparse attention methods developed for autoregressive LLMs~\citep{tangquest, zhang2023h2o, liu2023scissorhands, xiao2024efficient}. These methods select a subset of KV positions per query, reducing memory traffic. However, when naively adapting such methods to diffusion language models (DLMs), we encounter a \textbf{\kvinflation{} problem} (\fig{KVInflationTeaser}), as different queries within the same block tend to select different KV positions. Since attention remains memory-bound, the overhead is determined by the size of the \textbf{union} of KV positions of all queries in the block. Let $B$ be the block size, the size of this union can increase by at most $B$ times compared with per-query budget, effectively negating the intended speedup from sparsity.

To mitigate this problem, we observe that block-wise diffusion provides a unique structure absent in autoregressive decoding: \textbf{\LocalityRepChanges{}} across denoising steps. As shown in \fig{locality_intro}, between consecutive denoising iterations, only a small fraction of tokens undergo substantial updates to their hidden representations; we refer to these as \textbf{\Activetokens{}}. The remaining \textbf{\Stabletokens{}} have queries that change minimally, meaning their attention patterns, and therefore their attention outputs, remain approximately constant.

Building on this observation, we introduce \ours{} (\textbf{Lo}cality-aware \textbf{S}parse \textbf{A}ttention), a method that exploits \localityrepchanges{} to reduce KV cache memory operations. The key idea is to compute sparse prefix attention only for \activetokens{} at each denoising step, since they are the only queries whose attention patterns change substantially. For the remaining \stabletokens{} (i.e. tokens whose hidden representation has negligible change across denoising steps), we reuse cached prefix-attention outputs from the previous step. By reducing the number of queries that participate in sparse attention from the block size $B$ to $|\mathcal{A}|$ (where $\mathcal{A}$ is the set of \activetokens{}), we shrink the union of selected KV positions and directly lower memory traffic, leading to higher speedup. Beyond efficiency gains, \ours{} also improves accuracy compared to naive sparse attention. The key advantage is that for \stabletokens{}, our method preserves the full attention information from the cached prefix, whereas sparse attention can only access a subset of KV Cache. Therefore, for the majority of the tokens, \ours{} attention computation is more accurate, making the performance degradation less severe.

We summarize our contributions as follows:
\begin{itemize}[leftmargin=*,itemsep=2pt,topsep=2pt]
    \item 
    We observe \textbf{\LocalityRepChanges{}} in block-wise diffusion: across denoising steps, only a small fraction of tokens (\textbf{\Activetokens{}}) undergo substantial representation changes, while the majority (\textbf{\Stabletokens{}}) maintain approximately constant. Details in \sect{locality_rep_changes}.
    
    \item We identify the \textbf{\kvinflation{} problem} in block-wise DLM inference: when applying sparse attention to a block of $B$ queries, the attention kernel must load the \emph{union} of selected prefix KV positions across the block, making the speedup less than expected. Details in \sect{kv_inflation}.
    
    \item We propose \ours{}, a sparse attention method that reuses cached prefix attention outputs for \stabletokens{} and computes sparse attention only for \activetokens{}, reducing the union of selected KV indices by $\sim 1.5\times$ in practice. Details in \sect{method}.

    \item  We evaluate \ours{} on SDAR-8B~\cite{cheng2025sdar}, Trado-8B, and Trado-4B~\cite{wang2025revolutionizing} across LongBench and reasoning benchmarks. Compared to SparseD~\cite{wang2025sparsed}, Sparse-dLLM~\cite{song2026sparse}, and QUEST~\cite{tangquest}, \ours{} achieves up to +9 points accuracy improvement on LongBench at aggressive sparsity while maintaining $1.5\times$ lower attention density. On NVIDIA RTX A6000 GPUs, \ours{} delivers up to $4.14\times$ attention speedup, with the trend transferring to newer hardware ($3.67\times$ on RTX 5090).
\end{itemize}

\section{Preliminaries}
\lblsect{prelim}

\subsection{Block-Wise Diffusion Language Models}
\lblsect{block_wise_dlm}

Block-wise diffusion language models (DLMs) generate text in blocks of $B$ tokens, where each block is treated as a single generation unit. Blocks are generated autoregressively, while attention within each block is bidirectional.

At a denoising step, the queries, keys and values corresponding to the current block are denoted as $\mQ_b, \mK_b, \mV_b \in \R^{B\times d}$, where $d$ is the head dimension and the subscript $b$ stands for \emph{block}.
Since the denoising process follows an autoregressive manner, we maintain a KV cache of length $L$ to avoid recomputation. We denote the cached prefix keys/values as $\mK_p,\mV_p\in\R^{L\times d}$, where the subscript $p$ stands for \emph{prefix}.
Note that $\mK_p$ and $\mV_p$ are fixed across denoising steps, while $\mK_b$ and $\mV_b$ are updated at each step.

Consider a single attention head. The attention computation can be written as
\begin{align*}
    \mO
    &=
    \softmax\!\left(
    \frac{\mQ_b [\mK_p,\mK_b]^\top}{\sqrt{d}}
    \right)
    [\mV_p,\mV_b].
    \label{eq:full_attn}
\end{align*}

\subsection{Attention Decomposition With Online Softmax}

Starting from the full attention computation, we decompose the computation into a prefix part and a suffix part using the online-softmax method adopted in FlashAttention~\citep{dao2022flashattention}. We track each part's log-normalizer (log-sum-exp) and attention output first, then merge them together. To simplify notation, we consider a single attention head and assume the query $\vq \in \sR^d$ has only one token.

We define the score operator $\gS$ as
$\gS(\vq,\mK) = \frac{\vq \mK^\top}{\sqrt{d}} .$
Applying this operator to the prefix and current block yields score vectors
$\vs_p = \gS(\vq,\mK_p) \in \sR^L$ and $\vs_b = \gS(\vq,\mK_b) \in \sR^B$.
From these scores, we compute the corresponding log-normalizers $L$ (which is a scalar):
\[
L_p = \log \sum \exp(\vs_p),
\qquad
L_b = \log \sum \exp(\vs_b).
\]
The attention outputs computation can also be rewritten using the log-normalizers:
\begin{align*}
\vo_p = \softmax(\vs_p)\,\mV_p &= \frac{1}{e^{L_p}} \exp(\vs_p)\,\mV_p \in \sR^d, \\
\vo_b = \softmax(\vs_b)\,\mV_b &= \frac{1}{e^{L_b}} \exp(\vs_b)\,\mV_b \in \sR^d.
\end{align*}
We store the local log-normalizers $L_p$ and $L_b$, together with the attention outputs $\vo_p$ and $\vo_b$, then merge them using the online-softmax technique to obtain the final attention output.

When the query is a matrix $\mQ_b \in \sR^{B\times d}$, the size of the log-normalizers is $\sR^B$ and the attention outputs is $\sR^{B\times d}$, both are independent of $L$, meaning that the memory complexity is minimal when context length is long.

\subsection{Sparse Attention Algorithms for LLMs}
For LLMs, sparse attention methods approximate dense attention over the prefix by selecting a small subset of prefix positions.
Given query $\vq_i$, a sparse selector returns an index set $\mathcal{I}\subseteq\{1,\dots,L\}$ with budget $|\mathcal{I}|=k$, and the attention output is computed only over the selected keys/values. In this paper, we use the QUEST~\cite{tangquest} algorithm and naively adapted it for DLMs as a baseline to compare against. QUEST partitions the KV cache into contiguous pages of size $g$ and maintains lightweight min and max keys for each page. At inference time, it computes a query-dependent upper bound on the attention score for each page and selects the top $\frac{k}{g}$ pages for exact attention computation. Since the min and max keys are $g\times$ smaller than the original KV cache, the selection overhead is $\frac{g}{2}\times$ smaller than dense attention, which is minimal when the page size is large.
Note that \ours{} (introduced in \sect{method}) is largely orthogonal to the choice of sparse selector algorithm.

\begin{figure}[t]
\centering
\includegraphics[width=0.75\linewidth]{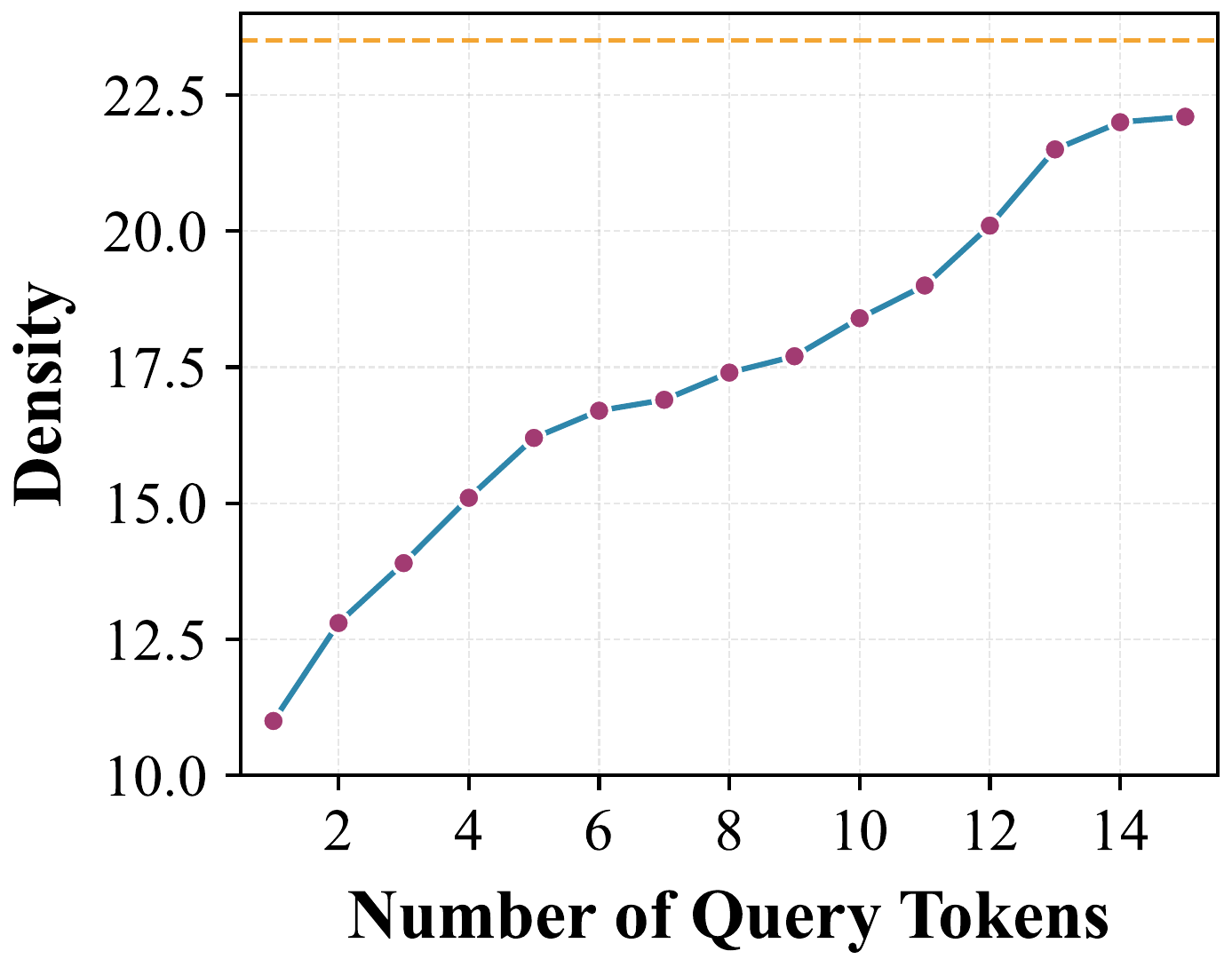}
\caption{
KV-cache load reduction with locality-aware sparse prefix attention.
X-axis: number of active query tokens in the block; Y-axis: percentage of the full KV cache that must be loaded. Evaluated on Trado-8B-Instruct on TriviaQA with block size 16 and 64K context length.
\ours{} only loads prefix KV positions in the union $\mathcal{I}=\bigcup_{i\in\mathcal{A}}\mathcal{I}_i$ for the \activetokens{} set $\mathcal{A}$, rather than the union over all $B$ queries.
The dashed orange line shows the KV load under QUEST, which must load the union across all queries.}
\label{fig:KVLoadReduction}
\end{figure}

\begin{figure}[t]
    \centering
    \includegraphics[width=0.8\linewidth]{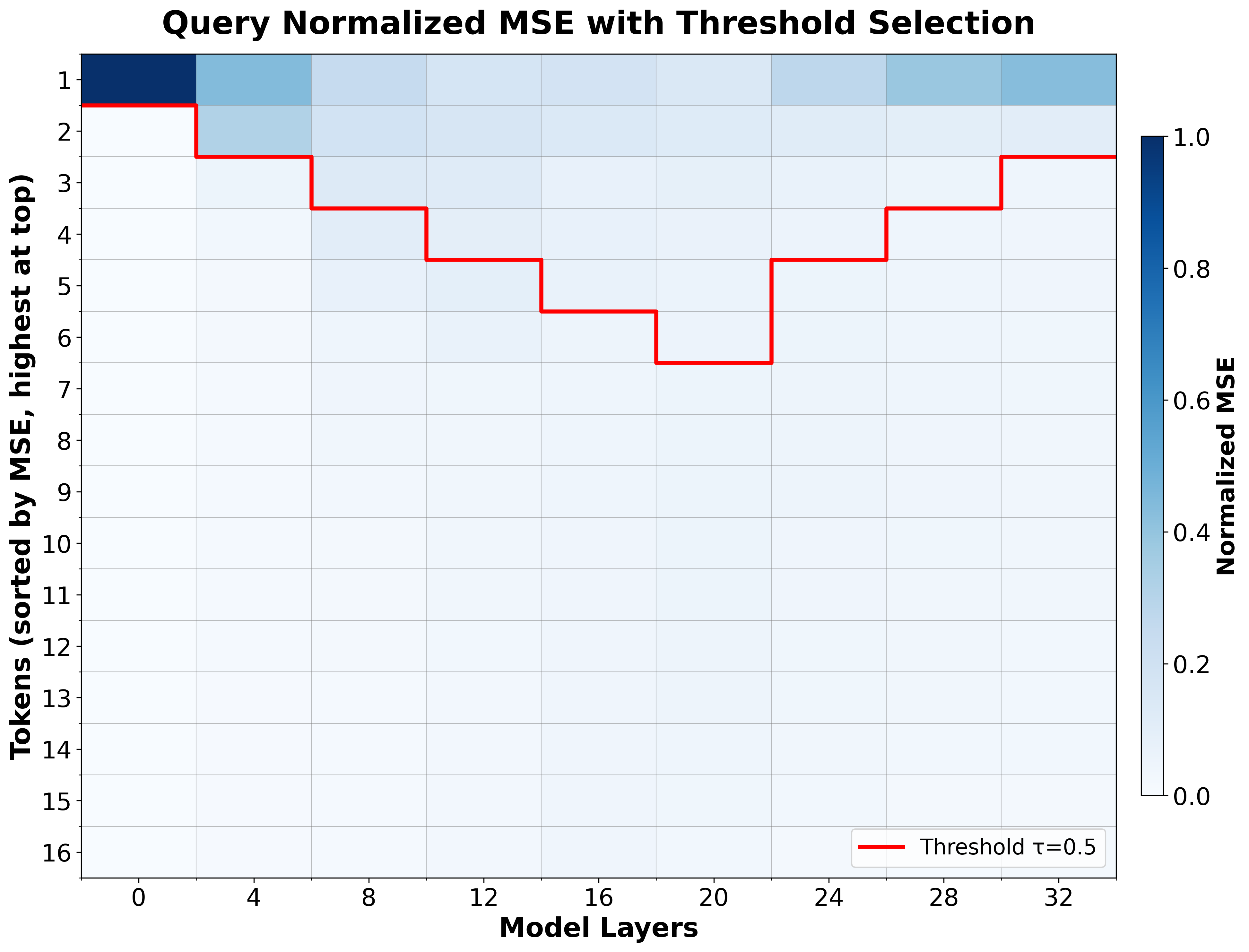}
    \caption{Visualizing locality across denoising steps. We plot the distribution of per-token MSE changes in queries between steps $t\!-\!1$ and $t$. For each layer, we sort the tokens in decreasing order of MSE changes from top to bottom in this heatmap. We observe that only a small fraction of tokens exhibit large changes (we show a threshold line which corresponds to 50\% of the total MSE change is exhibited by tokens above that line), motivating the reuse of cached-prefix attention for the stable tokens.}
    \label{fig:LayerLocality}
\end{figure}

\section{Motivation}
\lblsect{motivation}

\subsection{\kvinflation{} Problem}\lblsect{kv_inflation}
In autoregressive LLM decoding, sparse attention methods restrict each query to attend to only 
$k$ KV positions, making the attention latency roughly proportional to $k$. As a result, reducing the number of KV vectors loaded generally yields proportional speedup.

However, for DLMs, the situation is different. Suppose at each denoising step, the model processes a \emph{block} of $B$ query tokens. The attention workload consists of $B$ query tokens attending to $(L + B)$ KV cache tokens.
A naive adaptation of sparse attention applies a fixed budget $k$ \emph{per query token} in the block, 
producing index sets $\{\mathcal{I}_i\}_{i=1}^B$ with $|\mathcal{I}_i|=k$ for each query $i$.
To ensure all important information is captured, we compute the union of selected KV positions across all queries. The effective number of KV positions being loaded is denoted as $L^{\prime}$:
\begin{equation*}
L^{\prime} = \Big|\bigcup_{i=1}^{B} \mathcal{I}_i\Big|.
\label{eq:union_kv}
\end{equation*}

Since the block size $B$ is typically not large enough to make attention compute-bound, $L^{\prime}$ directly determines the memory traffic and thus the attention computation latency. The key insight is that loading a KV position incurs similar overhead whether it is used by one query or all queries in the block, making $L^{\prime}$ the critical factor for performance.

In this case, we identify a \textbf{\kvinflation{} problem} which makes sparse attention extremely inefficient for DLM workloads. Even though each query selects only selected $k$ KV positions, different queries may select different positions. Therefore, $L^{\prime}$ can become much larger than the intended budget $k$, negating the intended speedup from sparsity.
This phenomenon is well illustrated in \fig{KVInflationTeaser}.

In the worst case, $L^{\prime}$ can grow to $\min(L,\,Bk)$. This large union size increases memory traffic and leads to sub-optimal speedup performance. As indicated in \fig{KVLoadReduction}, the \kvinflation{} problem can lead to ${\sim}2.0\times$ higher attention density.

\begin{figure*}[t]
\centering
\includegraphics[
  width=0.99\linewidth,
]{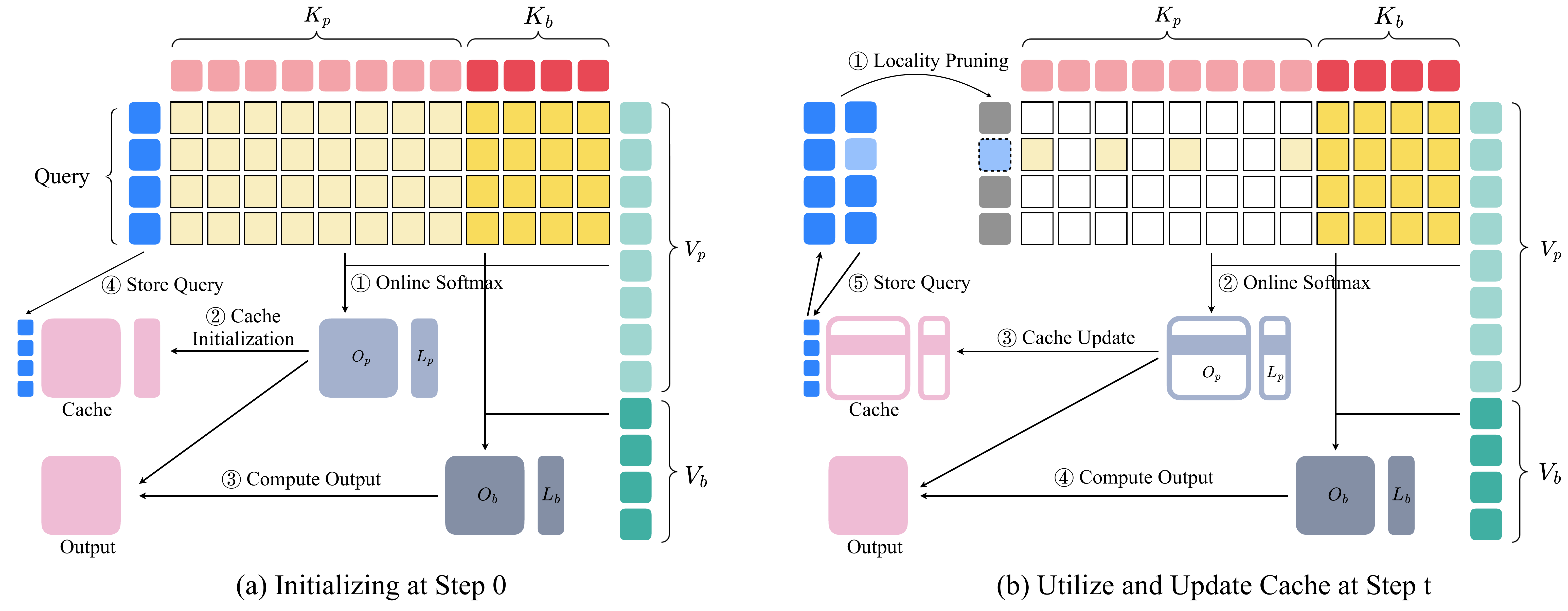}
\caption{
Overview of the locality-aware sparse prefix attention workflow.
We (i) measure per-token query changes and select the changed-token set $\mathcal{A}$, (ii) run QUEST to obtain per-token prefix indices and take their union $\mathcal{I}$, (iii) load prefix KV for $\mathcal{I}$ and compute updated prefix statistics for queries in $\mathcal{A}$ while reusing cached statistics for stable tokens, (iv) compute dense within-block attention, and (v) merge prefix and block contributions via online softmax and cache the updated prefix statistics for the next denoising step.
}
\label{fig:MethodOverview}
\end{figure*}

\subsection{\LocalityRepChanges{} Across Denoising Steps}\lblsect{locality_rep_changes}
Our solution begins by observing that block diffusion decoding offers an additional structure that autoregressive LLM decoding does not have: \emph{\localityrepchanges{} across denoising steps}.
Between two consecutive denoising steps $t\!-\!1$ and $t$, only a small subset of tokens in the current block are updated from [MASK] to actual tokens (see Figure~\ref{fig:locality_intro}). Tokens surrounding these updated positions experience significant changes in their hidden representations, while the remaining tokens' hidden states remain approximately constant.

Based on this observation, we categorize tokens into two groups: \textbf{\activetokens{}}, which undergo substantial changes in their hidden representations, and \textbf{\stabletokens{}}, which have approximately constant hidden states across consecutive denoising steps. We refer to this phenomenon as \textbf{\LocalityRepChanges{}}.

\paragraph{Quantifying locality for each token.}
To measure locality, we track the per-token change in queries, keys, and values using mean squared error (MSE).
For $\mX \in \{\mQ, \mK_b, \mV_b\}$, let $\vx^{(t)} \in \mathbb{R}^{B\times d}$ denote hidden states at denoising step $t$.
We define the locality score as the per-token MSE between the current and previous step.
\begin{equation}
\Delta_x^{(t)} = \operatorname{mean}\left(\frac{1}{d}
\left\lVert \vx^{(t)} - \vx^{(t-1)} \right\rVert_2^2\right)_{\text{token axis}} \in \sR^{B}.
\label{eq:mse_locality}
\end{equation}
We visualize the distribution of $\Delta_x^{(t)}$ across tokens in \fig{locality_intro}. The visualization confirms that \activetokens{} exhibit high locality scores, while \stabletokens{} have small scores. Notably, tokens at positions being decoded in the previous step show the largest changes.

We also observe that locality is layer-dependent. As shown in \fig{LayerLocality}, we sort the tokens by their locality scores and plot the distribution of locality scores for each layer. We observe that early layers and later layers tend to be more localized than middle layers.

\section{Methodology}
\lblsect{method}

In this section, we describe the methodology of our \ours{} method based on the aforementioned observations.

\subsection{Locality Pruning}

Let $\Delta \in \mathbb{R}^B$ denote the per-token locality score vector defined in \eqn{mse_locality}.
We rank tokens by their locality scores in descending order and select the Top-$k$ tokens with the largest changes, forming the set $\mathcal{A}$ of \textbf{\activetokens{}}.
Tokens not in $\mathcal{A}$ are regarded as \textbf{\stabletokens{}}.

\subsection{Locality-Aware Sparse Attention for Prefix Tokens}
We then propose a locality-aware sparse attention algorithm to reduce the KV cache loads. We cache and reuse the attention outputs for \stabletokens{}, and only compute the attention for \activetokens{} using sparse attention.
\paragraph{Reuse attention outputs for \stabletokens{}.}
For \stabletokens{} $i\notin\mathcal{A}$, since the query hidden states remain stable, the attention outputs for these tokens are also stable. Therefore, we can reuse the prefix-attention statistics $(\vo_p, L_p)$ from the previous step:
\begin{equation*}
(\vo_p, L_p) \leftarrow (\vo_p^{\mathrm{prev}}, L_p^{\mathrm{prev}}), \quad \text{token} \notin\mathcal{A}.
\label{eq:reuse_rule}
\end{equation*}

\providecommand{\avgnote}[1]{\raisebox{-0.35ex}{\textcolor{mydarkorange}{\scriptsize #1}}}
\begin{table*}[t]
    \centering
    \renewcommand{\arraystretch}{1.15}
    \setlength{\tabcolsep}{7pt}
    \resizebox{0.85\linewidth}{!}{%
    \begin{tabular}{c c | c c c c c l}
    \toprule
    \textbf{KV Cache} 
    & & \multicolumn{6}{c}{\textbf{LongBench Accuracy (Trado-8B-Instruct)}} \\

    \cmidrule(lr){3-8}
    
    \textbf{Per-Query Budget}
    & \textbf{Method}
    & \textbf{HotPotQA} 
    & \textbf{TriviaQA} 
    & \textbf{NarrativeQA} 
    & \textbf{Qasper} 
    & \textbf{MultiFieldQA} 
    & \textbf{Average}\avgnote{\,($\uparrow$)} \\
    
    \midrule

    \textbf{-}  & Dense 
    & 49.45\% & 84.79\% & 19.04\% & 17.75\% & 53.29\% & 44.86\% \\
    \midrule

    \multirow{2}{*}{128}  
      & \QUEST 
      & 29.17\% & 69.21\% & 7.46\%  & 13.64\% & 38.21\% & 31.54\% \\
      & \cellcolor{cyan!20}\ours{} 
      & \cellcolor{cyan!20}48.27\% 
      & \cellcolor{cyan!20}83.79\% 
      & \cellcolor{cyan!20}17.19\% 
      & \cellcolor{cyan!20}15.58\% 
      & \cellcolor{cyan!20}45.00\% 
      & \cellcolor{cyan!20}\textbf{41.97\%}\avgnote{(+10.43)} \\
    \cmidrule(l{0.8em}r{0.8em}){1-8}

    \multirow{2}{*}{256}  
      & \QUEST 
      & 32.95\% & 75.23\% & 8.50\%  & 14.04\% & 42.46\% & 34.64\% \\
      & \cellcolor{cyan!20}\ours{} 
      & \cellcolor{cyan!20}44.53\% 
      & \cellcolor{cyan!20}81.82\% 
      & \cellcolor{cyan!20}19.42\% 
      & \cellcolor{cyan!20}17.11\% 
      & \cellcolor{cyan!20}47.81\% 
      & \cellcolor{cyan!20}\textbf{42.14\%}\avgnote{(+7.50)} \\
    \cmidrule(l{0.8em}r{0.8em}){1-8}

    \multirow{2}{*}{512}  
      & \QUEST 
      & 34.58\% & 79.33\% & 8.99\%  & 16.57\% & 44.04\% & 36.70\% \\
      & \cellcolor{cyan!20}\ours{} 
      & \cellcolor{cyan!20}44.19\% 
      & \cellcolor{cyan!20}84.97\% 
      & \cellcolor{cyan!20}18.39\% 
      & \cellcolor{cyan!20}13.30\% 
      & \cellcolor{cyan!20}50.14\% 
      & \cellcolor{cyan!20}\textbf{42.20\%}\avgnote{(+5.50)} \\
    \cmidrule(l{0.8em}r{0.8em}){1-8}

    \multirow{2}{*}{1024} 
      & \QUEST 
      & 39.88\% & 78.84\% & 7.37\%  & 17.35\% & 45.00\% & 37.69\% \\
      & \cellcolor{cyan!20}\ours{} 
      & \cellcolor{cyan!20}48.45\% 
      & \cellcolor{cyan!20}82.32\% 
      & \cellcolor{cyan!20}18.89\% 
      & \cellcolor{cyan!20}15.78\% 
      & \cellcolor{cyan!20}48.94\% 
      & \cellcolor{cyan!20}\textbf{42.88\%}\avgnote{(+5.19)} \\
    \bottomrule
    \end{tabular}
    }
    \caption{
    Accuracy (\%) on LongBench under different retrieval budgets for \textbf{Trado-8B-Instruct}.
    \ours{} consistently outperforms \QUEST across all budget settings, with particularly large gains at low budgets (e.g., +10.43 at budget 128). The $\Delta$ shown is the improvement over \QUEST. Full comparison including SparseD is in \tbl{appendix_sparsed}.
    }
    \label{tab:qa_budget_results}
\end{table*}

\begin{table*}[t]
    \centering
    \renewcommand{\arraystretch}{1.15}
    \setlength{\tabcolsep}{7pt}
    \resizebox{0.8\linewidth}{!}{%
    \begin{tabular}{c c | c c c c c l}
    \toprule
    \textbf{KV Cache}
    & & \multicolumn{6}{c}{\textbf{LongBench KV-Cache Density}} \\

    \cmidrule(lr){3-8}

    \textbf{Per-Query Budget}
    & \textbf{Method}
    & \textbf{HotPotQA}
    & \textbf{TriviaQA}
    & \textbf{NarrativeQA}
    & \textbf{Qasper}
    & \textbf{MultiFieldQA}
    & \textbf{Average}\avgnote{\,($\downarrow$)} \\

    \midrule

    \multirow{2}{*}{128}  & \QUEST & 2.93\% & 3.80\% & 1.90\% & 6.72\% & 5.87\% & 4.24\% \\
     & \cellcolor{cyan!20}\ours{} & \cellcolor{cyan!20}1.71\% & \cellcolor{cyan!20}2.29\% & \cellcolor{cyan!20}0.98\% & \cellcolor{cyan!20}4.15\% & \cellcolor{cyan!20}3.77\% & \cellcolor{cyan!20}2.58\%\avgnote{($1.64\times$)} \\
    \cmidrule(l{0.8em}r{0.8em}){1-8}

    \multirow{2}{*}{256}  & \QUEST & 5.84\% & 7.07\% & 3.89\% & 12.47\% & 11.14\% & 8.08\% \\
     & \cellcolor{cyan!20}\ours{} & \cellcolor{cyan!20}3.37\% & \cellcolor{cyan!20}4.47\% & \cellcolor{cyan!20}2.05\% & \cellcolor{cyan!20}7.90\%  & \cellcolor{cyan!20}7.43\% & \cellcolor{cyan!20}5.04\%\avgnote{($1.60\times$)} \\
    \cmidrule(l{0.8em}r{0.8em}){1-8}

    \multirow{2}{*}{512}  & \QUEST & 10.74\% & 13.06\% & 7.24\% & 23.13\% & 19.71\% & 14.78\% \\
     & \cellcolor{cyan!20}\ours{} & \cellcolor{cyan!20}6.63\%  & \cellcolor{cyan!20}8.54\%  & \cellcolor{cyan!20}3.99\% & \cellcolor{cyan!20}15.39\% & \cellcolor{cyan!20}14.00\% & \cellcolor{cyan!20}9.71\%\avgnote{($1.52\times$)} \\
    \cmidrule(l{0.8em}r{0.8em}){1-8}

    \multirow{2}{*}{1024} & \QUEST & 19.10\% & 22.85\% & 13.06\% & 39.06\% & 34.21\% & 25.66\% \\
     & \cellcolor{cyan!20}\ours{} & \cellcolor{cyan!20}12.57\% & \cellcolor{cyan!20}16.24\% & \cellcolor{cyan!20}7.50\%  & \cellcolor{cyan!20}28.81\% & \cellcolor{cyan!20}26.04\% & \cellcolor{cyan!20}18.23\%\avgnote{($1.41\times$)} \\
    \bottomrule
    \end{tabular}
    }
    \caption{KV-cache density (percentage of prefix tokens selected) on LongBench under different budgets. On all datasets and budget configurations, \ours{} achieves an average of $1.54\times$ lower KV-cache density than the baseline.}
    \label{tab:qa_budget_density}
\end{table*}

\providecommand{\avgnote}[1]{\raisebox{-0.35ex}{\textcolor{mydarkorange}{\scriptsize #1}}}
\begin{table*}[t]
    \centering
    \renewcommand{\arraystretch}{1.15}
    \setlength{\tabcolsep}{7pt}
    \resizebox{0.85\linewidth}{!}{%
    \begin{tabular}{c c | c c c c c l}
    \toprule
    \textbf{KV Cache} 
    & & \multicolumn{6}{c}{\textbf{LongBench Accuracy (SDAR-8B-Chat)}} \\

    \cmidrule(lr){3-8}
    
    \textbf{Per-Query Budget}
    & \textbf{Method}
    & \textbf{HotPotQA} 
    & \textbf{TriviaQA} 
    & \textbf{NarrativeQA} 
    & \textbf{Qasper} 
    & \textbf{MultiFieldQA} 
    & \textbf{Average}\avgnote{\,($\uparrow$)} \\
    
    \midrule

    \textbf{-}  & Dense 
    & 49.35\% & 85.72\% & 19.06\% & 18.25\% & 49.49\% & 44.37\% \\
    \midrule

    \multirow{2}{*}{128}  
      & \QUEST 
      & 27.31\% & 70.40\% & 6.44\%  & 17.29\% & 41.40\% & 32.57\% \\
      & \cellcolor{cyan!20}\ours{} 
      & \cellcolor{cyan!20}43.36\% 
      & \cellcolor{cyan!20}80.32\% 
      & \cellcolor{cyan!20}18.69\% 
      & \cellcolor{cyan!20}15.63\% 
      & \cellcolor{cyan!20}46.74\% 
      & \cellcolor{cyan!20}\textbf{40.95\%}\avgnote{(+8.38)} \\
    \cmidrule(l{0.8em}r{0.8em}){1-8}

    \multirow{2}{*}{256}  
      & \QUEST 
      & 32.57\% & 78.36\% & 8.73\%  & 18.21\% & 43.20\% & 36.21\% \\
      & \cellcolor{cyan!20}\ours{} 
      & \cellcolor{cyan!20}45.68\% 
      & \cellcolor{cyan!20}83.16\% 
      & \cellcolor{cyan!20}15.65\% 
      & \cellcolor{cyan!20}13.17\% 
      & \cellcolor{cyan!20}48.33\% 
      & \cellcolor{cyan!20}\textbf{41.20\%}\avgnote{(+4.99)} \\
    \cmidrule(l{0.8em}r{0.8em}){1-8}

    \multirow{2}{*}{512}  
      & \QUEST 
      & 32.50\% & 78.43\% & 9.80\%  & 19.36\% & 43.08\% & 36.63\% \\
      & \cellcolor{cyan!20}\ours{} 
      & \cellcolor{cyan!20}47.77\% 
      & \cellcolor{cyan!20}83.29\% 
      & \cellcolor{cyan!20}17.37\% 
      & \cellcolor{cyan!20}14.12\% 
      & \cellcolor{cyan!20}49.90\% 
      & \cellcolor{cyan!20}\textbf{42.49\%}\avgnote{(+5.86)} \\
    \cmidrule(l{0.8em}r{0.8em}){1-8}

    \multirow{2}{*}{1024} 
      & \QUEST 
      & 32.59\% & 80.92\% & 11.00\% & 19.28\% & 42.30\% & 37.22\% \\
      & \cellcolor{cyan!20}\ours{} 
      & \cellcolor{cyan!20}47.84\% 
      & \cellcolor{cyan!20}85.93\% 
      & \cellcolor{cyan!20}18.44\% 
      & \cellcolor{cyan!20}14.70\% 
      & \cellcolor{cyan!20}48.43\% 
      & \cellcolor{cyan!20}\textbf{43.07\%}\avgnote{(+5.85)} \\
    \bottomrule
    \end{tabular}
    }
    \caption{
    Accuracy (\%) on LongBench under different retrieval budgets for \textbf{SDAR-8B-Chat}.
    Results are consistent with Trado-8B (\tbl{qa_budget_results}): \ours{} consistently outperforms \QUEST, with the largest gains at low budgets (+8.38 at budget 128). Full comparison including SparseD is in \tbl{appendix_sparsed}.
    }
    \label{tab:sdar_longbench}
\end{table*}

\paragraph{Sparse attention with smaller union size for \activetokens{}.}
For \activetokens{} $i\in\mathcal{A}$, we apply the same sparse selector (QUEST) with budget $k$ to select the prefix indices, but \textit{only for the \activetokens{} in $\mathcal{A}$}:
\begin{equation*}
\mathcal{I}_i = \mathrm{QUEST}(\vq_i, \mK_p; k), \quad i\in\mathcal{A},
\quad
\mathcal{I} = \bigcup_{i\in\mathcal{A}} \mathcal{I}_i.
\label{eq:quest_union}
\end{equation*}

We load the prefix key/value vectors indexed by the union set $\mathcal{I}$.
Then, for all queries in $\mathcal{A}$, we compute prefix attention over this shared set $\mathcal{I}$. 
After the computation, we update the prefix statistics $(\vo_p, L_p)$ for these tokens:
\begin{equation*}
    (\vo_p, L_p) \leftarrow (\vo_p^{\mathrm{current}}, L_p^{\mathrm{current}}), \quad \text{token} \in \mathcal{A}.
    \label{eq:update_rule}
\end{equation*}

\paragraph{Initialization Workflow.}
For the first denoising iteration for a block, cached statistics are unavailable; we fall back to dense prefix attention for initialization and start to apply the locality-aware sparse attention from the second iteration onward. We find that this initialization step is crucial for the performance of \ours{}. The reason is that the dense prefix attention is more accurate than the sparse prefix attention, and the stable tokens can benefit from the cached attention outputs by capturing all information in the prefix.

\subsection{Computation of the Suffix Tokens and Final Output}
We next focus on the computation of $\mK_b$ and $\mV_b$. Although keys and values also exhibit the same locality structure as queries, their relevant attention computation is minimal since $B\ll L$. Therefore, we compute the attention between $Q_b$, $K_b$, and $V_b$ densely, and get the low-dimensional statistics $(\vo_b, L_b)$ for suffix tokens.

Finally, we merge the prefix and current block contributions using the online-softmax merge to obtain the final attention output.
We cache the updated prefix statistics $(\vo_p, L_p)$, together with the query states for the next denoising step.

\subsection{Why Locality Reduces KV Loads: Union Size}
The KV-cache traffic for sparse prefix attention is governed by the number of unique prefix positions (or KV tiles) that must be fetched:
$|\mathcal{I}| = \left|\bigcup_{i\in\mathcal{A}}\mathcal{I}_i\right|$.
Compared to the naive per-token sparse scheme (which uses $\mathcal{A}=\{1,\dots,B\}$), our locality-aware scheme uses a smaller set of queries, which can only reduce (never increase) the union:
\begin{equation*}
\left|\bigcup_{i\in\mathcal{A}}\mathcal{I}_i\right|
\;\le\;
\left|\bigcup_{i=1}^{B}\mathcal{I}_i\right|
\;\le\;
\min(L,\,|\mathcal{A}|k).
\end{equation*}
The reduction is generally \emph{not} proportional to $|\mathcal{A}|$: it depends on the overlap structure among the $\{\mathcal{I}_i\}$.
Nevertheless, shrinking the participating query set from $B$ to $|\mathcal{A}|$ eliminates many query-induced selections and typically decreases the number of KV cache that must be loaded.
As shown in \fig{KVLoadReduction}, a smaller union size leads to smaller KV cache loads and higher speedup.

\subsection{Efficiency Analysis}
The overhead introduced by \ours{} consists of three main components: (1) locality pruning to identify \activetokens{}, (2) sparse selection to select the prefix indices that participate in the sparse attention computation and compute its union, and (3) computing the final attention output.

The locality pruning step requires calculating the per-token locality scores and ranking tokens by their scores, which incurs $\mathcal{O}(B \times d + B \log B)$ FLOPs overhead. The sparse selector incurs $\mathcal{O}(\frac{B}{g} \times d)$ overhead, where $g$ is the page size used by QUEST. Computing their union incurs $\mathcal{O}(\frac{B}{g})$ overhead. The final attention output computation also incurs $\mathcal{O}(B \times d)$ overhead, as the online-softmax merge is applied to LSE and local attention outputs. The total overhead is $\mathcal{O}(B \times d + \frac{B}{g} \times d)$. Since $B \ll L$ in typical DLM settings, these overheads are minimal compared to the actual attention computation.

\paragraph{Memory overhead.}
The additional memory required by \ours{} is minimal. We cache the prefix attention output $\vo_p \in \mathbb{R}^{B \times d}$ and the log-normalizer $L_p \in \mathbb{R}^{B}$ for each attention head, yielding $B(d+1)$ extra scalars per head. For typical settings ($B=16$, $d=128$), this amounts to ${\sim}2$\,KB per head in FP16, which is negligible compared to the KV cache size $\mathcal{O}(L \times d)$. The remaining memory traffic is dominated by loading the union of selected KV positions, which \ours{} directly reduces.

\providecommand{\avgnote}[1]{\raisebox{-0.35ex}{\textcolor{mydarkorange}{\scriptsize #1}}}
\begin{table*}[t]
    \centering
    \renewcommand{\arraystretch}{1.15}
    \setlength{\tabcolsep}{7pt}
    \resizebox{0.95\linewidth}{!}{%
    \begin{tabular}{c c | c c c c c l || c c c c c l}
    \toprule
    & & \multicolumn{6}{c||}{\textbf{Block Size = 16}} & \multicolumn{6}{c}{\textbf{Block Size = 32}} \\
    \cmidrule(lr){3-8} \cmidrule(lr){9-14}
    \textbf{Budget}
    & \textbf{Method}
    & \textbf{HotPotQA} 
    & \textbf{TriviaQA} 
    & \textbf{NarrativeQA} 
    & \textbf{Qasper} 
    & \textbf{MFieldQA} 
    & \textbf{Avg.}\avgnote{\,($\uparrow$)}
    & \textbf{HotPotQA} 
    & \textbf{TriviaQA} 
    & \textbf{NarrativeQA} 
    & \textbf{Qasper} 
    & \textbf{MFieldQA} 
    & \textbf{Avg.}\avgnote{\,($\uparrow$)} \\
    \midrule
    \textbf{-}  & Dense 
    & 38.16\% & 75.61\% & 18.80\% & 29.57\% & 35.42\% & 39.51\%
    & 36.22\% & 75.23\% & 19.58\% & 13.40\% & 31.52\% & 35.19\% \\
    \cmidrule(lr){1-14}
    \multirow{2}{*}{128}
      & \QUEST 
      & 30.01\% & 69.14\% & 9.83\%  & 26.05\% & 30.79\% & 33.16\%
      & 25.55\% & 70.83\% & 9.53\%  & 11.20\% & 27.30\% & 28.88\% \\
      & \cellcolor{cyan!20}\ours{}
      & \cellcolor{cyan!20}33.65\% 
      & \cellcolor{cyan!20}66.46\% 
      & \cellcolor{cyan!20}14.78\% 
      & \cellcolor{cyan!20}24.71\% 
      & \cellcolor{cyan!20}30.07\% 
      & \cellcolor{cyan!20}\textbf{33.93\%}
      & \cellcolor{cyan!20}28.62\% 
      & \cellcolor{cyan!20}69.50\% 
      & \cellcolor{cyan!20}12.15\% 
      & \cellcolor{cyan!20}13.78\% 
      & \cellcolor{cyan!20}28.25\% 
      & \cellcolor{cyan!20}\textbf{30.46\%} \\
    \cmidrule(lr){1-14}
    \multirow{2}{*}{256}
      & \QUEST 
      & 33.06\% & 73.49\% & 12.34\% & 29.25\% & 34.67\% & 36.56\%
      & 31.61\% & 72.10\% & 10.87\% & 12.85\% & 29.10\% & 31.31\% \\
      & \cellcolor{cyan!20}\ours{}
      & \cellcolor{cyan!20}36.33\% 
      & \cellcolor{cyan!20}72.74\% 
      & \cellcolor{cyan!20}15.99\% 
      & \cellcolor{cyan!20}28.11\% 
      & \cellcolor{cyan!20}34.27\% 
      & \cellcolor{cyan!20}\textbf{37.49\%}
      & \cellcolor{cyan!20}32.14\% 
      & \cellcolor{cyan!20}72.20\% 
      & \cellcolor{cyan!20}15.99\% 
      & \cellcolor{cyan!20}14.80\% 
      & \cellcolor{cyan!20}28.68\% 
      & \cellcolor{cyan!20}\textbf{32.76\%} \\
    \bottomrule
    \end{tabular}%
    }
    \caption{LongBench accuracy (\%) for \textbf{Trado-4B-Instruct} under two block sizes. \ours{} consistently outperforms \QUEST across model sizes and block-size configurations, confirming that the locality-aware approach generalizes beyond the 8B scale.}
    \label{tab:trado4b_longbench}
\end{table*}

\begin{table}[t]
    \centering
    \small
    \setlength{\tabcolsep}{4pt}
    \resizebox{0.9\columnwidth}{!}{%
    \begin{tabular}{c c c c c c}
    \toprule
    \textbf{Budget} & \textbf{Method} & \textbf{HellaSwag} & \textbf{WinoGrande} & \textbf{BoolQ} & \textbf{Avg.} \\
    \midrule
    \textbf{--}  & Dense          & 69.60 & 63.60 & 72.92 & 68.71 \\
    \cmidrule(lr){1-6}
    \multirow{2}{*}{128}  & \QUEST        & 67.60 & 66.80 & 73.96 & 69.45 \\
     & \ours         & 70.80 & 66.00 & 72.92 & 69.91 \\
    \cmidrule(lr){1-6}
    \multirow{2}{*}{256}  & \QUEST        & 73.60 & 63.20 & 75.00 & 70.60 \\
     & \ours         & 70.40 & 65.20 & 73.96 & 69.85 \\
    \bottomrule
    \end{tabular}%
    }
    \caption{Accuracy (\%) on three commonsense reasoning benchmarks for Trado-8B-Instruct. \ours{} performs comparably to \QUEST{}; the slight gap at budget 256 is within statistical variance on these short-context tasks ($<$1K tokens) where the \kvinflation{} problem is less severe.
    }
    \label{tab:trador_results_transposed}
    \end{table}

\begin{figure*}[t]
    \centering
    \includegraphics[width=0.98\textwidth]{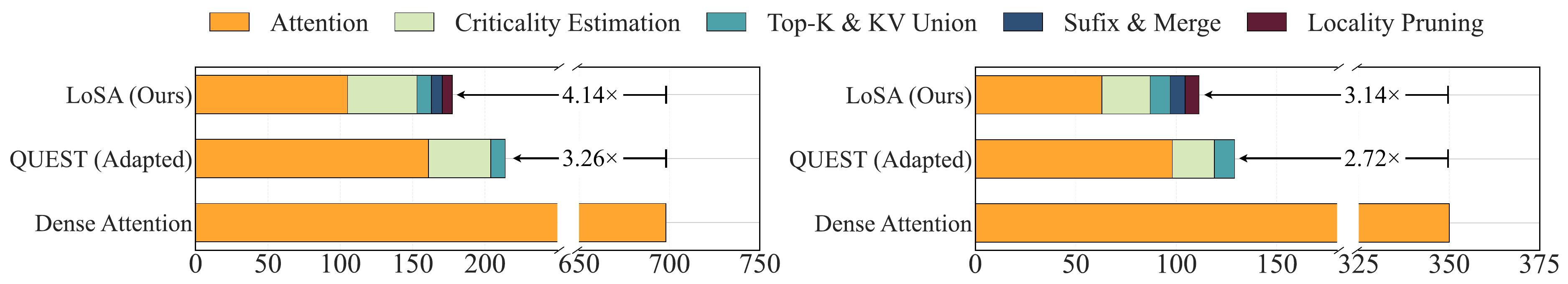}
    \caption{Latency breakdown comparison on prefix attention using Trado-8B-Instruct on TriviaQA with RTX A6000. \textbf{Left:} 64K context length with block size 16. \textbf{Right:} 32K context length with block size 32. \ours{} and QUEST (Adapted) achieve $4.14\times$ and $3.26\times$ speedup over Dense Attention, respectively, by reducing memory-bound attention operations through sparsity. \ours{} further improves over QUEST through locality-aware reuse.}
    \label{fig:speedup}
\end{figure*}

\section{Experiments and Results}

\subsection{Experimental Settings}
\label{sec:settings}

\textbf{Models.} We evaluate \ours{} on three block-wise diffusion language models: Trado-8B-Instruct, Trado-4B-Instruct~\citep{wang2025revolutionizing}, and SDAR-8B-Instruct~\cite{cheng2025sdar}. All models follow the semi-autoregressive block-wise generation paradigm. We set the block size to 16 if not specified.

\textbf{Baselines.} We compare against \QUEST{}~\cite{tangquest}, a per-query sparse attention method originally designed for autoregressive LLMs, which we adapt for DLMs by applying it independently to each query in a block. We additionally compare with SparseD~\cite{wang2025sparsed} in \tbl{appendix_sparsed}.

\textbf{Datasets.} We evaluate mainly on LongBench~\cite{bai2024longbench} to examine their long-context ability. To be specific, we evaluate on HotPotQA~\cite{yang2018hotpotqa}, TriviaQA~\cite{joshi2017triviaqa}, NarrativeQA~\cite{kovcisky2018narrativeqa}, Qasper~\cite{dasigi2021dataset}, MultiFieldQA~\cite{bai2024longbench}, and their average. We also evaluate on commonsense reasoning benchmarks, including HellaSwag~\cite{zellers2019hellaswag}, WinoGrande~\cite{sakaguchi2021winogrande}, and BoolQ~\cite{clark2019boolq}.

\paragraph{Implementation}
We implement our algorithm using customized CUDA and Triton~\cite{tillet2019triton} kernels, together with attention kernels from FlashInfer~\cite{ye2025flashinfer}. We pick the top-5 query tokens in our locality pruning algorithm.

\subsection{LongBench Results}

We evaluate \ours{} on LongBench across multiple models. As shown in \tbl{qa_budget_results}, on Trado-8B-Instruct, \ours{} consistently outperforms \QUEST{} across all budget configurations and datasets. At budget 128, \ours{} achieves 41.97\% average accuracy, outperforming \QUEST{} (31.54\%) by +10.43. At budget 256, \ours{} (42.14\%) leads \QUEST{} (34.64\%) by +7.50. At higher budgets (512 and 1024), both methods converge toward dense accuracy, with \ours{} maintaining a clear advantage. Notably, \ours{} maintains accuracy close to the Dense baseline even at aggressive sparsity levels, demonstrating that locality-aware reuse effectively preserves important attention information.
In terms of attention density, \ours{} maintains an average of $1.54\times$ lower attention density than \QUEST{} across all budget configurations and datasets (\tbl{qa_budget_density}).

On SDAR-8B-Chat (\tbl{sdar_longbench}), the trends are consistent: \ours{} outperforms \QUEST{} at all budget levels, with the largest gains at budget 128 (+8.38). At higher budgets (512, 1024), both methods converge toward dense accuracy.

To verify that \ours{} generalizes across model sizes, we further evaluate on Trado-4B-Instruct with both block size 16 and block size 32 (\tbl{trado4b_longbench}). \ours{} consistently outperforms \QUEST{} across both configurations, confirming that the locality-aware approach is not specific to one model size or block-size setting.

\subsection{Commonsense Reasoning Results}

We evaluate \ours{} on commonsense reasoning benchmarks using Trado-8B-Instruct. \tbl{trador_results_transposed} shows accuracy results under different budgets. \ours{} achieves competitive or better accuracy compared to \QUEST{} across all three datasets. At budget 128, \ours{} achieves an average accuracy of $69.91\%$, slightly outperforming \QUEST{}'s 69.45\%. At budget 256, \QUEST{} achieves $70.60\%$ average accuracy while \ours{} maintains $69.85\%$; this slight gap is within statistical variance and is expected since these short-context benchmarks (typically $<$1K tokens) do not strongly exhibit the \kvinflation{} problem that \ours{} is designed to address. \ours{}'s advantage is most pronounced in long-context settings (\tbl{qa_budget_results}).
\subsection{Latency Analysis}

We measure end-to-end latency of prefix attention computation to validate that our method translates sparsity into wall-clock speedup. We evaluate two configurations: (1) 64K context length with 16 token block size, and (2) 32K context length with 32 token block size, both using Trado-8B-Instruct and testing on TriviaQA.

\fig{speedup} compares the latency breakdown of \ours{}, \QUEST{}, and Dense Attention. The latency breakdown consists of several components: (1) \textbf{Criticality Estimation}, which refers to the sparse attention selector (QUEST) that identifies relevant KV positions for each query; (2) \textbf{Locality Pruning}, which identifies \activetokens{} by computing locality scores based on representation changes; (3) \textbf{Cache Management}, which stores and retrieves cached attention outputs for \stabletokens{}; and (4) \textbf{Attention Computation}, which performs the actual sparse attention operations over selected KV positions.

On the RTX A6000, \ours{} achieves a $4.14\times$ speedup over Dense Attention, outperforming \QUEST{}'s $3.26\times$ speedup. The key advantage of \ours{} is that locality-aware reuse substantially reduces the number of queries requiring fresh attention computation. By caching and reusing attention outputs for \stabletokens{}, \ours{} minimizes memory-bound operations while maintaining accuracy, leading to lower attention computation time despite the additional overhead of locality pruning and cache management.

\paragraph{Scalability to newer GPUs.}
We additionally measure latency on an RTX 5090 under the 64K context length, block size 16 setting. Dense attention takes $650\,\mu$s, while \ours{} completes in $177\,\mu$s ($3.67\times$ speedup) and \QUEST{} in $228\,\mu$s ($2.85\times$ speedup). The speedup trend transfers across GPU generations because the workflow remains memory-bound: the dominant cost is KV data movement rather than compute. As memory bandwidth improves on newer hardware, kernel launch overhead and other fixed costs become a larger relative fraction, leading to slightly lower but still substantial speedups.

\section{Related Work}

\subsection{Diffusion Language Models}

Diffusion models have achieved remarkable success in continuous domains such as images and audio, inspiring efforts to adapt them for discrete text generation, which enables any-order decoding~\citep{kang2025parallelbench}. Early work explored discrete diffusion through absorbing states~\citep{austin2021structured} and multinomial diffusion processes. More recently, masked diffusion language models (MDLM)~\citep{sahoo2024simple} demonstrated that simple masked diffusion with modern architectures can match autoregressive baselines on language modeling benchmarks. SEDD~\citep{lou2024discrete} introduced score entropy-based training for discrete diffusion, achieving state-of-the-art perplexity. LLaDA~\citep{nie2025large} and Dream~\citep{ye2025dream} scaled masked diffusion models demonstrating competitive performance with autoregressive models of similar scale.

\textbf{Block-wise diffusion models.} Recent work has explored semi-autoregressive diffusion that generates tokens in blocks~\cite{kang2025parallelbench}. BD3-LMs~\citep{arriola2025block} interpolate between fully autoregressive and fully any-order generation by producing fixed-size blocks autoregressively while using diffusion within each block. SDAR~\citep{cheng2025sdar} converts an autoregressive LLM into a block-diffusion model. Trado~\citep{wang2025revolutionizing} applies advanced post-training techniques to enhance their reasoning capabilities.
Efforts to extend DLMs to longer contexts include LongLLaDA~\citep{liu2026longllada}, which unlocks long-context capabilities in diffusion LLMs, and UltraLLaDA~\citep{he2025ultrallada}, which scales context length to 128K tokens.
Our work is designed for this block-wise setting.

\textbf{Efficient inference for DLMs.}
Several concurrent works address inference acceleration in DLMs.
Fast-dLLM~\citep{wu2025fastdllm} and its successor Fast-dLLM v2~\citep{wu2025fastdllmv2} enable KV caching and any-order decoding for diffusion LLMs without additional training.
SparseD~\citep{wang2025sparsed} observes that sparse attention patterns change little across denoising steps and reuses a fixed sparse pattern computed at initialization.
Sparse-dLLM~\citep{song2026sparse} accelerates diffusion LLMs by dynamically evicting KV cache entries.
Our method differs fundamentally from these approaches: LoSA reuses \emph{attention outputs} (not patterns) for stable tokens, allowing them to retain full dense-attention information, and applies dynamic sparse attention only to active tokens whose representations have changed.

\subsection{Sparse Attention for Long-Context LLMs}

Reducing the quadratic complexity of attention has been extensively studied. Fixed sparse patterns such as local windows combined with global tokens~\citep{child2019generating, beltagy2020longformer, zaheer2020bigbird} reduce complexity to linear but require architectural modifications during training. Content-based sparse attention methods like Reformer~\citep{kitaev2020reformer} use locality-sensitive hashing to identify relevant keys. Linear attention variants~\citep{katharopoulos2020transformers, choromanski2021rethinking} approximate softmax attention with kernel feature maps, enabling recurrent computation.

For pretrained LLMs at inference time, dynamic sparse attention methods select relevant KV positions per query without retraining. QUEST~\citep{tangquest} groups KV cache into pages and selects pages based on estimated attention scores. H\textsubscript{2}O~\citep{zhang2023h2o} maintains a dynamic cache of ``heavy hitter'' tokens that accumulate high attention mass. Scissorhands~\citep{liu2023scissorhands} exploits the persistence of important tokens across generation steps. StreamingLLM~\citep{xiao2024efficient} discovers that keeping initial ``sink'' tokens enables stable streaming inference. SnapKV~\citep{li2024snapkv} and PyramidKV~\citep{cai2024pyramidkv} compress the KV cache based on attention patterns observed during prefill. MInference~\citep{jiang2024minference} accelerates long-context prefilling through dynamic sparse patterns.


\subsection{KV Cache Optimization and Efficient Serving}

Beyond sparse attention, several orthogonal techniques improve KV cache efficiency. FlashAttention~\citep{dao2022flashattention, dao2023flashattention2} optimizes memory access patterns through tiling, reducing memory traffic for dense attention without approximation. PagedAttention~\citep{kwon2023efficient} enables non-contiguous KV cache storage, improving memory utilization in serving systems. SageAttention~\cite{zhang2025sageattention,zhang2024sageattention2,zhang2025sageattention3,zhang2025sageattention2++} quantizes the attention computation to lower precision to speedup attention further.

\section{Limitations}
While \ours{} is effective for long-context block-wise DLM inference, its advantage diminishes for short-context inputs (e.g., $<$1K tokens) where the \kvinflation{} problem is less severe and all methods perform comparably. Additionally, our current evaluation focuses on batch size 1; integrating \ours{} with batched serving systems (e.g., vLLM, TensorRT-LLM) is an interesting direction for future work. Finally, \ours{} requires an initial dense attention pass at the first denoising step of each block, which adds a fixed overhead that becomes amortized over subsequent denoising iterations.

\section{Conclusion}

In this paper, we identify the \kvinflation{} problem in block-wise diffusion language models as the main reason why naive sparse attention fails to accelerate block-wise DLMs. We observe \localityrepchanges{} across denoising steps and propose \ours{}, which reuses cached attention for \stabletokens{} and computes sparse attention only for \activetokens{}. This reduces the number of queries participating in sparse KV selection, effectively mitigating the \kvinflation{} problem. Across Trado-8B, SDAR-8B, and Trado-4B, \ours{} consistently outperforms SparseD, Sparse-dLLM, and \QUEST{} on LongBench, with up to +9.01 average accuracy improvement at aggressive sparsity (budget 128). \ours{} also achieves $1.54\times$ lower attention density and up to $4.14\times$ speedup on A6000 GPUs, with the speedup trend transferring to newer hardware (3.67$\times$ on RTX 5090).


\section*{Acknowledgements}
We acknowledge the gracious support from the FuriosaAI, Intel, Apple, NVIDIA, Macronix, and Mozilla team.
Furthermore, we appreciate the support from
Google Cloud, the Google TRC team Prof.~David Patterson, along with support from Google Gemini team, and Divy Thakkar.
Prof.~Keutzer's lab is also sponsored by funding through BDD and BAIR.
We also acknowledge support by the Director, Office of Science, Office of Advanced Scientific Computing Research, of the U.S. Department of Energy under Contract No. DE-AC02-05CH11231.
MWM would also like to acknowledge DARPA, DOE, NSF, and ONR.
DOE SciGPT grant. Our conclusions do not necessarily reflect the position or the policy of our sponsors, and no official endorsement should be~inferred.

\bibliography{icml}
\bibliographystyle{icml2026}

\newpage
\appendix
\onecolumn

\section{Appendix}

\subsection{Full LongBench Comparison with SparseD}
\providecommand{\avgnote}[1]{\raisebox{-0.35ex}{\textcolor{mydarkorange}{\scriptsize #1}}}
\begin{table*}[h]
    \centering
    \renewcommand{\arraystretch}{1.15}
    \setlength{\tabcolsep}{7pt}
    \resizebox{0.85\linewidth}{!}{%
    \begin{tabular}{c c | c c c c c l}
    \toprule
    \textbf{KV Cache} 
    & & \multicolumn{6}{c}{\textbf{LongBench Accuracy -- Comparison with SparseD}} \\

    \cmidrule(lr){3-8}
    
    \textbf{Per-Query Budget}
    & \textbf{Method}
    & \textbf{HotPotQA} 
    & \textbf{TriviaQA} 
    & \textbf{NarrativeQA} 
    & \textbf{Qasper} 
    & \textbf{MultiFieldQA} 
    & \textbf{Average}\avgnote{\,($\uparrow$)} \\
    
    \midrule
    \multicolumn{8}{c}{\textbf{Trado-8B-Instruct}} \\
    \midrule

    \textbf{-}  & Dense 
    & 49.45\% & 84.79\% & 19.04\% & 17.75\% & 53.29\% & 44.86\% \\
    \midrule

    \multirow{3}{*}{128}  
      & \QUEST 
      & 29.17\% & 69.21\% & 7.46\%  & 13.64\% & 38.21\% & 31.54\% \\
      & SparseD
      & 38.97\% & 64.64\% & 13.70\% & 7.95\%  & 39.55\% & 32.96\% \\
      & \cellcolor{cyan!20}\ours{} 
      & \cellcolor{cyan!20}48.27\% 
      & \cellcolor{cyan!20}83.79\% 
      & \cellcolor{cyan!20}17.19\% 
      & \cellcolor{cyan!20}15.58\% 
      & \cellcolor{cyan!20}45.00\% 
      & \cellcolor{cyan!20}\textbf{41.97\%}\avgnote{(+9.01)} \\
    \cmidrule(l{0.8em}r{0.8em}){1-8}

    \multirow{3}{*}{256}  
      & \QUEST 
      & 32.95\% & 75.23\% & 8.50\%  & 14.04\% & 42.46\% & 34.64\% \\
      & SparseD
      & 43.84\% & 73.63\% & 15.33\% & 13.26\% & 49.44\% & 39.10\% \\
      & \cellcolor{cyan!20}\ours{} 
      & \cellcolor{cyan!20}44.53\% 
      & \cellcolor{cyan!20}81.82\% 
      & \cellcolor{cyan!20}19.42\% 
      & \cellcolor{cyan!20}17.11\% 
      & \cellcolor{cyan!20}47.81\% 
      & \cellcolor{cyan!20}\textbf{42.14\%}\avgnote{(+3.04)} \\
    \cmidrule(l{0.8em}r{0.8em}){1-8}

    \multirow{3}{*}{512}  
      & \QUEST 
      & 34.58\% & 79.33\% & 8.99\%  & 16.57\% & 44.04\% & 36.70\% \\
      & SparseD
      & 47.84\% & 80.78\% & 14.08\% & 13.85\% & 52.16\% & 41.74\% \\
      & \cellcolor{cyan!20}\ours{} 
      & \cellcolor{cyan!20}44.19\% 
      & \cellcolor{cyan!20}84.97\% 
      & \cellcolor{cyan!20}18.39\% 
      & \cellcolor{cyan!20}13.30\% 
      & \cellcolor{cyan!20}50.14\% 
      & \cellcolor{cyan!20}\textbf{42.20\%}\avgnote{(+0.46)} \\
    \cmidrule(l{0.8em}r{0.8em}){1-8}

    \multirow{3}{*}{1024} 
      & \QUEST 
      & 39.88\% & 78.84\% & 7.37\%  & 17.35\% & 45.00\% & 37.69\% \\
      & SparseD
      & 48.98\% & 80.25\% & 19.43\% & 15.77\% & 50.11\% & 42.91\% \\
      & \cellcolor{cyan!20}\ours{} 
      & \cellcolor{cyan!20}48.45\% 
      & \cellcolor{cyan!20}82.32\% 
      & \cellcolor{cyan!20}18.89\% 
      & \cellcolor{cyan!20}15.78\% 
      & \cellcolor{cyan!20}48.94\% 
      & \cellcolor{cyan!20}42.88\% \\

    \midrule
    \multicolumn{8}{c}{\textbf{SDAR-8B-Chat}} \\
    \midrule

    \textbf{-}  & Dense 
    & 49.35\% & 85.72\% & 19.06\% & 18.25\% & 49.49\% & 44.37\% \\
    \midrule

    \multirow{3}{*}{128}  
      & \QUEST 
      & 27.31\% & 70.40\% & 6.44\%  & 17.29\% & 41.40\% & 32.57\% \\
      & SparseD
      & 42.09\% & 63.08\% & 12.94\% & 9.04\%  & 36.24\% & 32.68\% \\
      & \cellcolor{cyan!20}\ours{} 
      & \cellcolor{cyan!20}43.36\% 
      & \cellcolor{cyan!20}80.32\% 
      & \cellcolor{cyan!20}18.69\% 
      & \cellcolor{cyan!20}15.63\% 
      & \cellcolor{cyan!20}46.74\% 
      & \cellcolor{cyan!20}\textbf{40.95\%}\avgnote{(+8.27)} \\
    \cmidrule(l{0.8em}r{0.8em}){1-8}

    \multirow{3}{*}{256}  
      & \QUEST 
      & 32.57\% & 78.36\% & 8.73\%  & 18.21\% & 43.20\% & 36.21\% \\
      & SparseD
      & 48.39\% & 73.28\% & 15.83\% & 14.84\% & 47.37\% & 39.94\% \\
      & \cellcolor{cyan!20}\ours{} 
      & \cellcolor{cyan!20}45.68\% 
      & \cellcolor{cyan!20}83.16\% 
      & \cellcolor{cyan!20}15.65\% 
      & \cellcolor{cyan!20}13.17\% 
      & \cellcolor{cyan!20}48.33\% 
      & \cellcolor{cyan!20}\textbf{41.20\%}\avgnote{(+1.26)} \\
    \cmidrule(l{0.8em}r{0.8em}){1-8}

    \multirow{3}{*}{512}  
      & \QUEST 
      & 32.50\% & 78.43\% & 9.80\%  & 19.36\% & 43.08\% & 36.63\% \\
      & SparseD
      & 50.52\% & 81.83\% & 15.44\% & 18.52\% & 48.93\% & 43.05\% \\
      & \cellcolor{cyan!20}\ours{} 
      & \cellcolor{cyan!20}47.77\% 
      & \cellcolor{cyan!20}83.29\% 
      & \cellcolor{cyan!20}17.37\% 
      & \cellcolor{cyan!20}14.12\% 
      & \cellcolor{cyan!20}49.90\% 
      & \cellcolor{cyan!20}42.49\% \\
    \cmidrule(l{0.8em}r{0.8em}){1-8}

    \multirow{3}{*}{1024} 
      & \QUEST 
      & 32.59\% & 80.92\% & 11.00\% & 19.28\% & 42.30\% & 37.22\% \\
      & SparseD
      & 48.25\% & 83.44\% & 15.60\% & 16.15\% & 48.97\% & 42.48\% \\
      & \cellcolor{cyan!20}\ours{} 
      & \cellcolor{cyan!20}47.84\% 
      & \cellcolor{cyan!20}85.93\% 
      & \cellcolor{cyan!20}18.44\% 
      & \cellcolor{cyan!20}14.70\% 
      & \cellcolor{cyan!20}48.43\% 
      & \cellcolor{cyan!20}\textbf{43.07\%}\avgnote{(+0.59)} \\
    \bottomrule
    \end{tabular}
    }
    \caption{
    Full LongBench accuracy comparison including SparseD. The $\Delta$ shown is the improvement of \ours{} over the best competing method at each budget. At budget 1024 on Trado-8B and budget 512 on SDAR-8B, SparseD slightly outperforms \ours{} in average accuracy.
    }
    \label{tab:appendix_sparsed}
\end{table*}

\begin{figure*}[t]
\centering

\includegraphics[
  width=\textwidth,
  trim = 0mm 0mm 0mm 0mm,
  clip
]{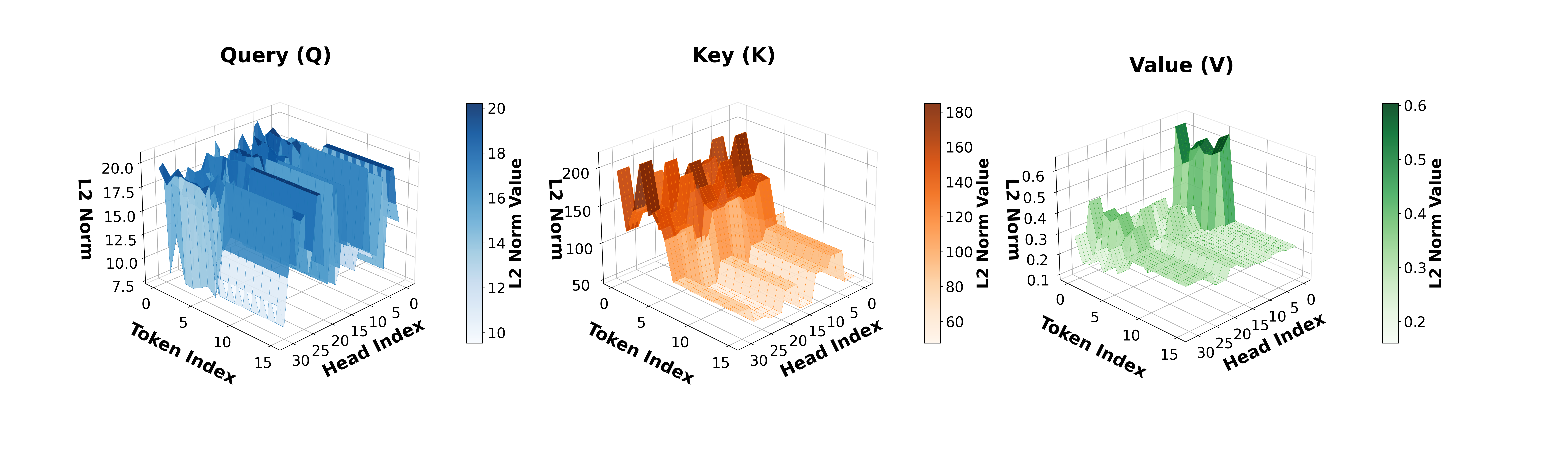}

\vspace{6pt}

\includegraphics[
  width=\textwidth,
  trim = 0mm 0mm 0mm 0mm,
  clip
]{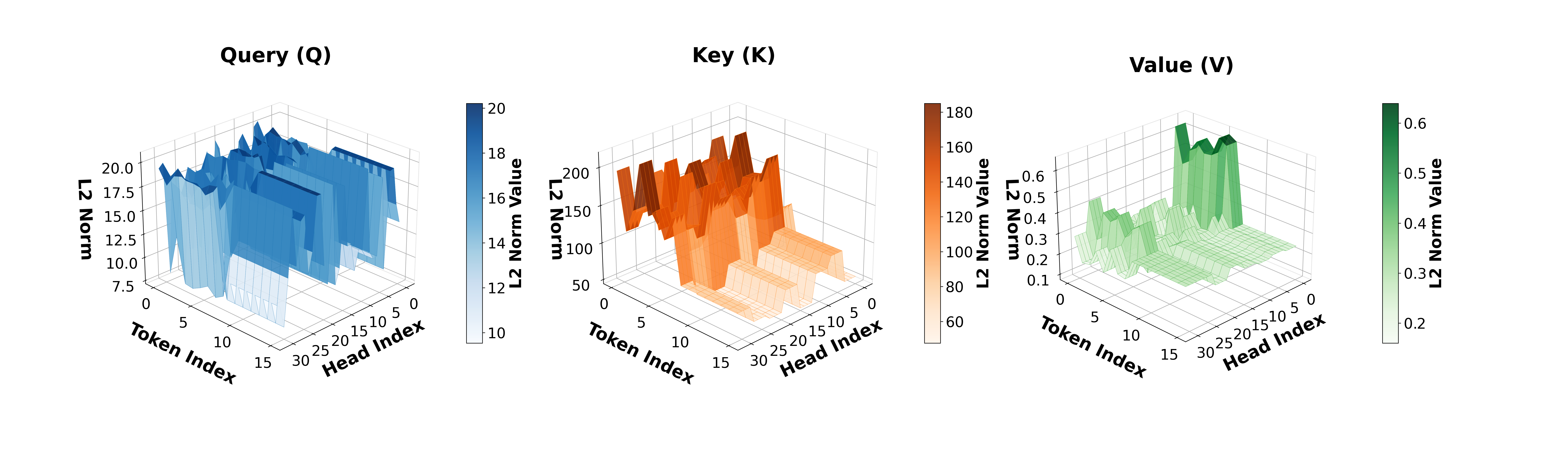}

\vspace{6pt}

\includegraphics[
  width=\textwidth,
  trim = 0mm 0mm 0mm 0mm,
  clip
]{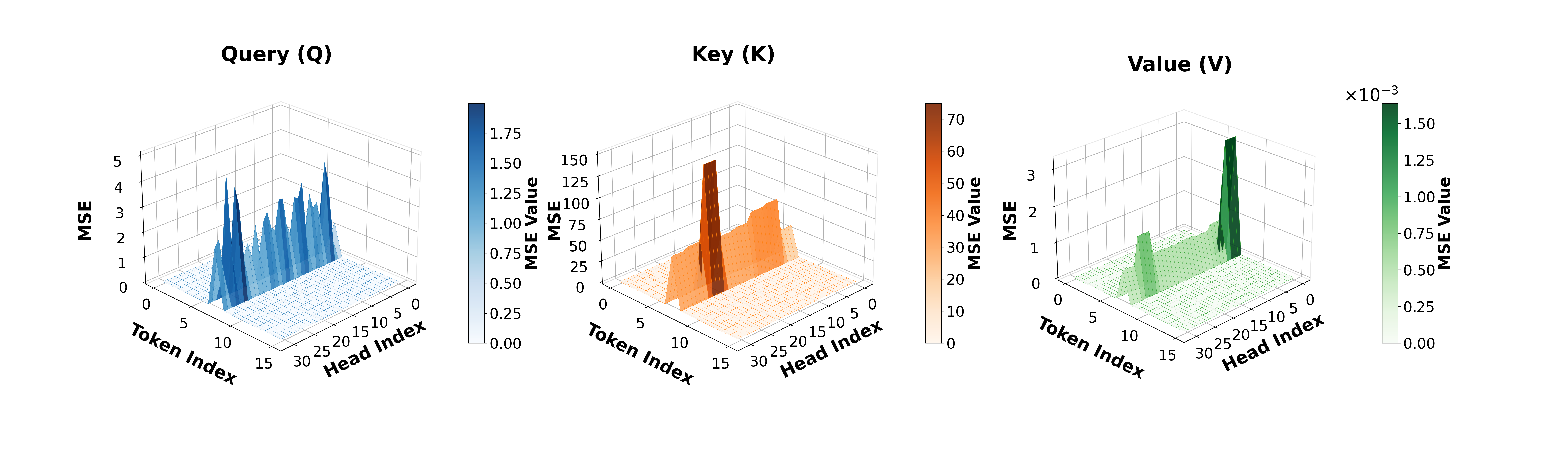}

\caption{
Visualizing representation change locality across denoising steps for query, keys and values of the token in the block that is being decoded.
\textbf{(Top)} L2 norm of vectors at step $t\!-\!1$.
\textbf{(Middle)} L2 norm of vectors at step $t$.
\textbf{(Bottom)} Per-token change in representations, measured using MSE between steps $t\!-\!1$ and $t$.
Only a small fraction of tokens exhibit large changes, motivating reuse of cached-prefix attention for stable tokens.
}
\label{fig:locality_qkv_appendix}
\end{figure*}


\begin{figure}[htbp]
    \centering
    \includegraphics[width=\textwidth]{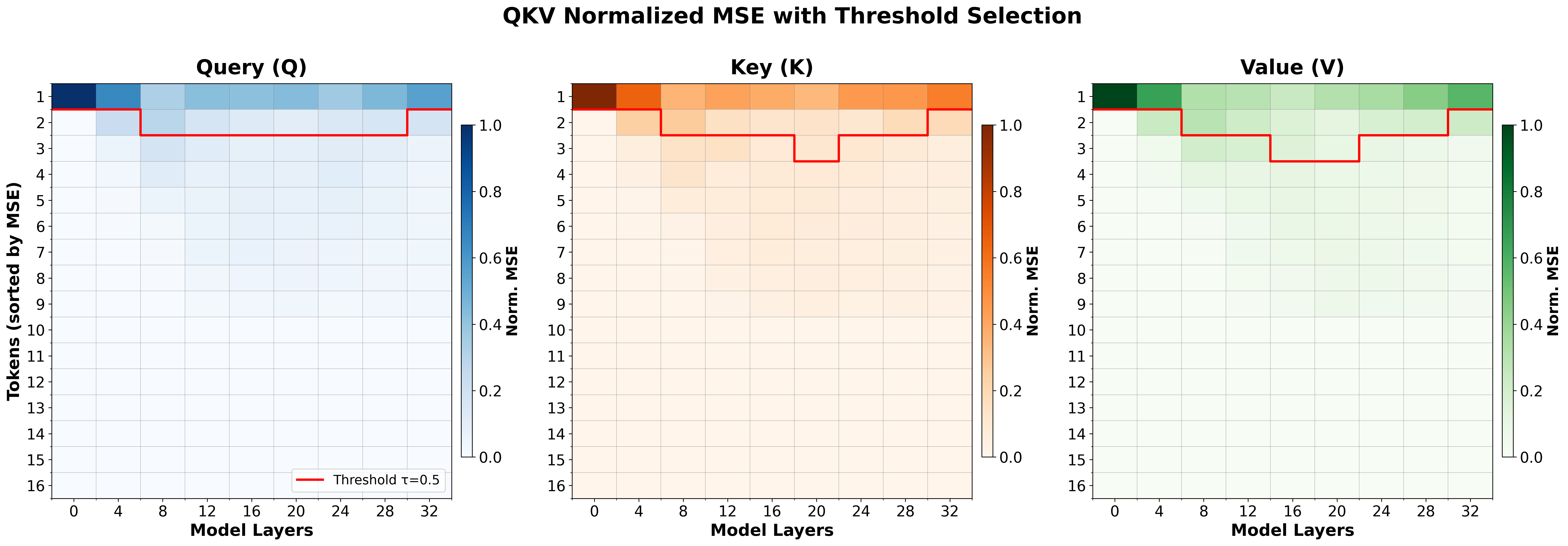}
    \caption{Combined QKV Normalized MSE Heatmap with Threshold Selection ($\tau=0.5$) for Data Sample 0. The red line indicates the boundary where cumulative normalized MSE reaches 50\% of the layer total. Tokens above the line are selected; tokens below are dimmed.}
    \label{fig:qkv_threshold_data0}
\end{figure}

\begin{figure}[htbp]
    \centering
    \includegraphics[width=\textwidth]{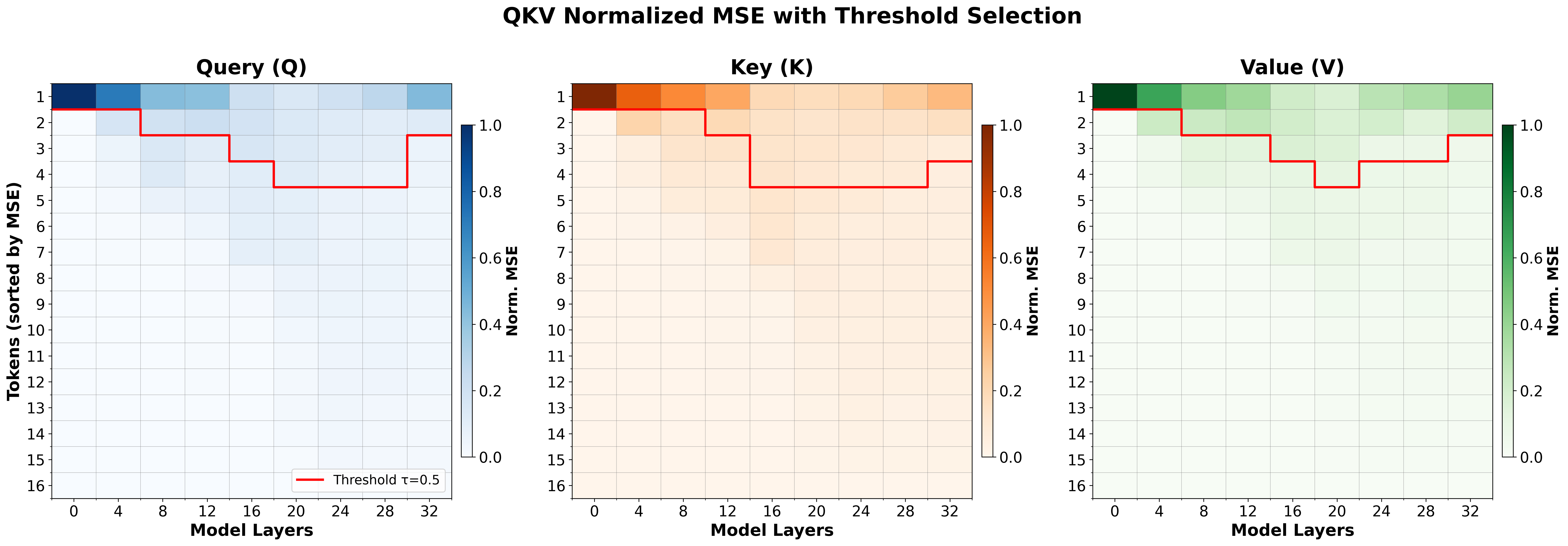}
    \caption{Combined QKV Normalized MSE Heatmap with Threshold Selection ($\tau=0.5$) for Data Sample 100.}
    \label{fig:qkv_threshold_data100}
\end{figure}

\begin{figure}[htbp]
    \centering
    \includegraphics[width=\textwidth]{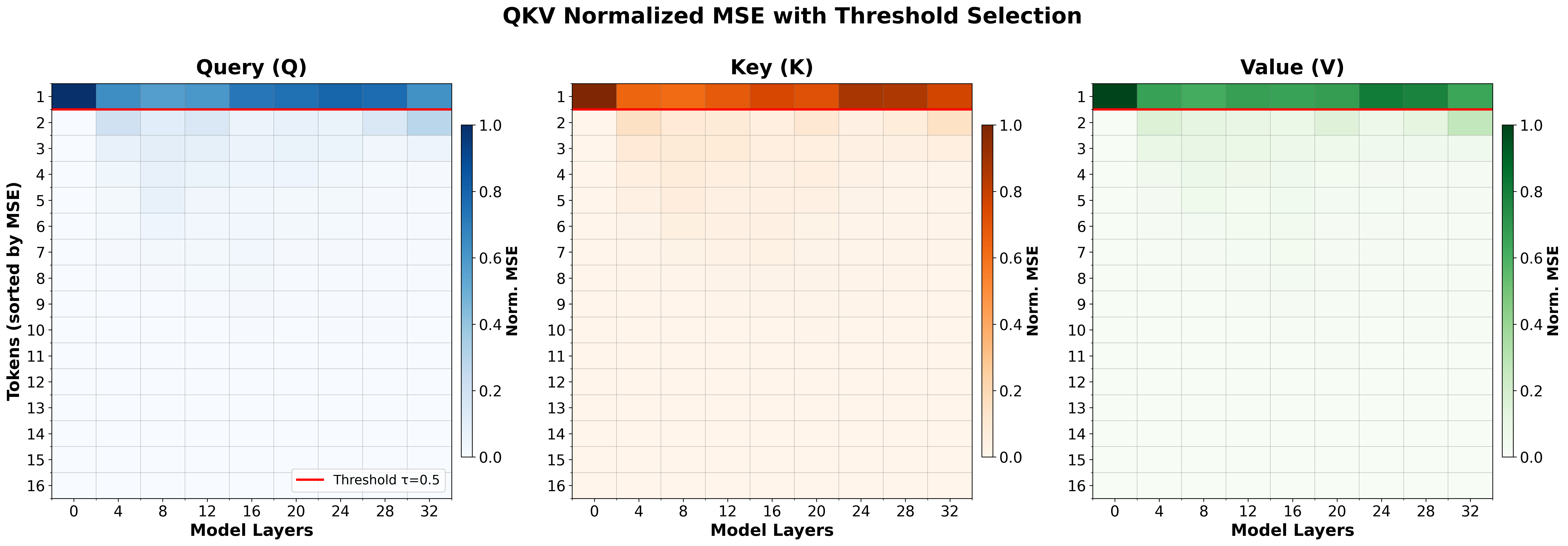}
    \caption{Combined QKV Normalized MSE Heatmap with Threshold Selection ($\tau=0.5$) for Data Sample 200.}
    \label{fig:qkv_threshold_data200}
\end{figure}

\begin{figure}[htbp]
    \centering
    \includegraphics[width=\textwidth]{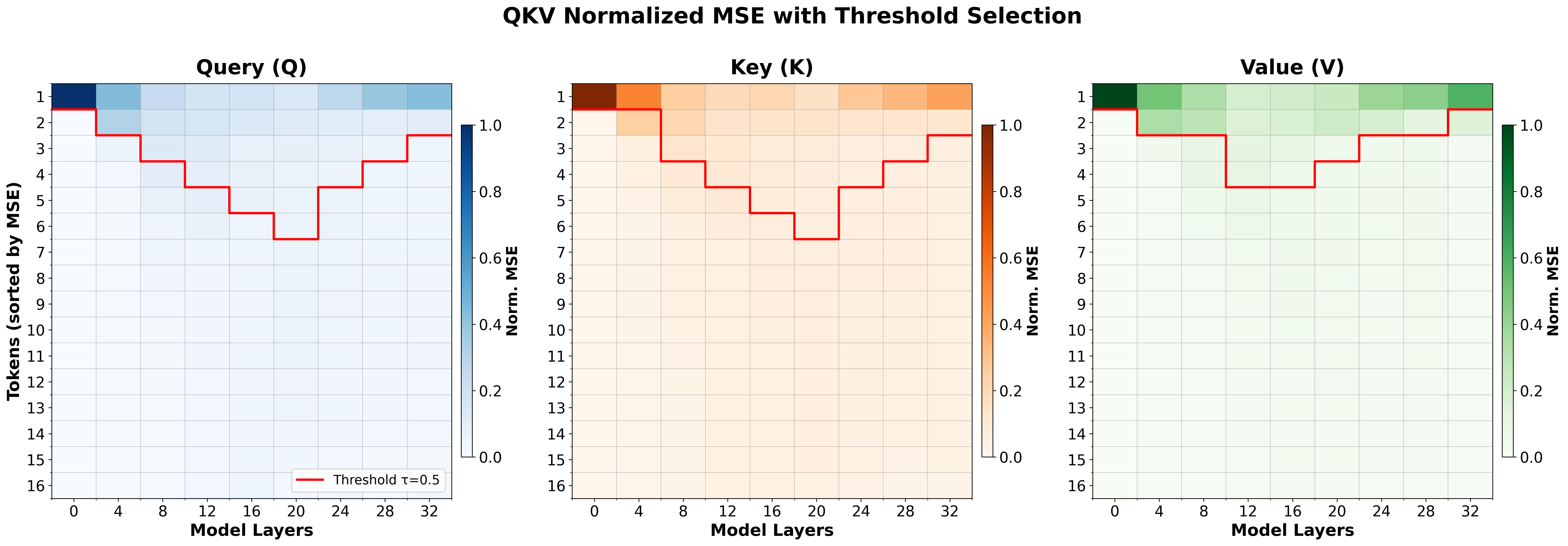}
    \caption{Combined QKV Normalized MSE Heatmap with Threshold Selection ($\tau=0.5$) for Data Sample 300.}
    \label{fig:qkv_threshold_data300}
\end{figure}

\end{document}